\documentclass[tikz,fleqn,12pt]{wlscirep}
\usepackage[english]{babel}
\usepackage[autostyle]{csquotes}
\usepackage[nameinlink]{cleveref}
\usepackage[utf8]{inputenc}
\usepackage[T1]{fontenc}
\usepackage{bm}
\usepackage{tikz}
\usepackage{hyperref}
\usepackage[nobiblatex]{xurl}
\usepackage{amsmath}
\usepackage{mathrsfs}
\usepackage{mathptmx}
\usepackage{float}
\usepackage{graphicx}
\usepackage{subcaption}
\usepackage{nameref}
\DeclareMathAlphabet{\altmathcal}{OMS}{cmsy}{m}{n}
\newcommand{\overbar}[1]{\mkern 1.5mu\overline{\mkern-1.5mu#1\mkern-1.5mu}\mkern 1.5mu}
\usetikzlibrary{arrows.meta,positioning,fit}

\MakeOuterQuote{"}
\MakeAutoQuote{«}{»}
\title{SIR-RL: Reinforcement Learning for Optimized Policy Control during Epidemiological Outbreaks in Emerging Market and Developing Economies}
\author[1]{Maeghal Jain}
\author[1]{Ziya Uddin}
\author[2,*]{Wubshet Ibrahim}
\affil[1]{SoET, BML Munjal University, Gurugram, Haryana, 122413, India}
\affil[2]{Department of Mathematics, Ambo University, Ambo, Ethiopia}
\affil[*]{e-mail: wubshet.ibrahim@ambou.edu.et}
\begin{abstract}
The outbreak of COVID-19 has highlighted the intricate interplay between public health and economic stability on a global scale. This study proposes a novel reinforcement learning framework designed to optimize health and economic outcomes during pandemics. The framework leverages the SIR model, integrating both lockdown measures (via a stringency index) and vaccination strategies to simulate disease dynamics. The stringency index, indicative of the severity of lockdown measures, influences both the spread of the disease and the economic health of a country. Developing nations, which bear a disproportionate economic burden under stringent lockdowns, are the primary focus of our study. By implementing reinforcement learning, we aim to optimize governmental responses and strike a balance between the competing costs associated with public health and economic stability. This approach also enhances transparency in governmental decision-making by establishing a well-defined reward function for the reinforcement learning agent. In essence, this study introduces an innovative and ethical strategy to navigate the challenge of balancing public health and economic stability amidst infectious disease outbreaks.
\end{abstract}
\begin{document}
\flushbottom
\maketitle
\thispagestyle{empty}
\section{Introduction}
In the past, global spread of infectious diseases was largely due to colonization, slavery, and war, leading to widespread illness and death from diseases like tuberculosis, polio, smallpox, and diphtheria. Medical advancements, better access to health care, and improved sanitation have worked towards improving the situation of mortality and morbidity linked to infectious diseases in the past twenty years. However, in low and lower-middle income countries the burden of infectious diseases still persists. The rapid pace of urbanization in low and middle-income countries, along with the rise in populations living in crowded, poor-quality homes, has led to new conditions that favor the emergence of infectious diseases~\cite{Baker2022,inequitableworld}.

Recently, the COVID-19 pandemic caused havoc worldwide. To date there have been 772 million cases and more than 6 million deaths~\cite{WHOCovid}. The pandemic triggered the sharpest economic recession in modern history with a 3\% decline, much worse than during the 2008-09 financial crisis~\cite{IMFCovid}. As nations grappled with the immediate health crisis, the economic fallout disproportionately affected vulnerable populations and exacerbated existing inequalities. Lockdowns and restrictions imposed to curb the spread of the virus led to widespread unemployment, business closures, and disruptions in global supply chains~\cite{Nicola2020}. The challenges faced by low and lower-middle income countries were particularly acute, highlighting the intricate interplay between public health and economic stability on a global scale~\cite{Gagnon2023}. 

The need for a nuanced understanding of how interventions impact both health outcomes and economic indicators became increasingly evident, prompting a comprehensive examination by epidemiologists to assist policymakers~\cite{Anderson2020}. The outbreak of COVID-19 has prompted epidemiologists to research on various aspects, including mobility control~\cite{Song2022,Chinazzi2020}, vaccination strategies~\cite{Nguyen2021, Kim2022}, non-pharmaceutical interventions (NPIs) like restricting population movements and gatherings, closing schools and businesses, requiring masks indoors~\cite{Jalloh2022,Caldwell2021,Ferguson2020}, and financial considerations~\cite{DeFoo2023}. Despite the numerous studies conducted, very few explore how common interventions meet multiple policy objectives or how a precise articulation of the main policy goals directs the selection of the most effective interventions in terms of health and economic results~\cite{Hollingsworth2011, Song2022,Pangallo2023,Ash2022,Ohi2020, PADMANABHAN2021102676,NBERw26981,analytical_covid_lockdown_model}. The economic impact of the COVID-19 pandemic varied between rich and poor countries. Although COVID-19 deaths had a slightly larger negative effect on the Gross Domestic Product (GDP) in advanced economies, this difference was not statistically significant. However, lockdown restrictions were found to have a more damaging impact on economic activity in emerging and developing economies~\cite{Gagnon2023,Redlin2022,Liang2021}. It's also suggested that an increase in COVID-19 cases was associated with the introduction of harsher NPIs and lockdown measures could be relaxed once vaccination rates increase~\cite{Redlin2022,Patel2021}.

Many economists have studied the effect of COVID-19 on the economy of nations~\cite{Gagnon2023,GagnonKorea,Deb2020,Eichenbaum2021}. In advanced economies like Korea, where the stringency index was below the median, the recession was milder than other advanced economies like the United Kingdom where the stringency was much higher~\cite{GagnonKorea}, they achieved it mostly with very aggressive testing, contact tracing, and enforced quarantines~\cite{Lim2023,KoreaMinister}. In India, social distancing and containment measures have been effective in reducing the number of COVID-19 cases but have come with economic costs. Social distancing had the most adverse effect on the economy in areas with high urbanization~\cite{Deb2020}.

In this paper, we optimize the government policies regarding stringency as it controls both the spread of the disease and the economy. To model the epidemiological data~\cite{WorldometerCorona} we use the simple SIR model without vital dynamics~\cite{Hethcote1989, Hethcote2008, ALLEN2017128}, as it is assumed that the timescale is small enough that it can be neglected~\cite{Cooper2020}. By lesioning the model, as opposed to proposing a new mathematical model with more specialized compartments to more accurately represent the actual environment~\cite{Bjrnstad2020, Mwalili2020}, we effectively model the disease progression. Our model (SIR with lockdown and time-varying vaccination rate) builds on the foundational SIR model, by accounting for the recovery reached through vaccination~\cite{Marinov2022,MaurciodeCarvalho2023,Thater,Turkyilmazoglu2022,YALADANDA2022101052} and the effects of lockdown~\cite{Hale2021,SIRLockdown,NBERw26981,NBERw26867}. Although the traditional SIR model is a valuable tool for understanding the spread of infectious diseases, it assumes that parameters like the transmission rate ($\beta$) and recovery rate ($\gamma$) are constant over time, which may not always be the case. In this paper, we propose a more sophisticated approach by introducing a time-dependent SIR model~\cite{TimeDependentSIR}, enabling us to account for the changing dynamics of the pandemic due to factors such as lockdowns and vaccination rates. This proposed model effectively addresses the real-world conditions acts as a solution that is both effective and extendable. However, the study has limitations; First, the deterministic SIR model (predecessor to our proposed model) fails to account for chance in disease spread and lacks confidence intervals on results and while stochastic models incorporate chance, they are typically more challenging to analyze than their deterministic counterparts~\cite{Hethcote2008}; Secondly, the underreporting of cases during the period selected by our study; Lastly, the reinforcement learning agent should be resistant to how the vaccination rate changes and different values for $\beta$ and $\gamma$ -- keeping them the same scopes the environment for succumbing to wishful thinking which can be potentially dangerous. Therefore, before an actual deployment of the model, it would be a good measure to introduce stochasticity to these parameters ($\beta$ and $\gamma$) and the vaccination rate ($\nu$). 

After modelling the disease with lockdown (via stringency index) and vaccination, we try to understand the effects of lockdown on the GDP~\cite{StringencyEconomicDecline,Cilloni2020,Arinaminpathy2010}. Therefore, decisions made by the government regarding the level of lockdown to be enforced plays a role on both the public health outcomes and economic stability during a pandemic. On one hand, stringent lockdown measures can effectively slow the spread of the disease, thereby improving public health outcomes. However, these measures often come at the cost of significant economic disruption, leading to job losses, business closures, and reduced economic growth. On the other hand, relaxing lockdown measures may help to mitigate the economic impact of the pandemic, but could result in increased disease transmission and worsened public health outcomes. In order to capture competing costs within the environment and achieve a balance between health and economic outcomes, we intend to employ reinforcement learning~\cite{Nguyen2022,Bastani2021,Song2022,Ohi2020,PADMANABHAN2021102676}. Not only does the formulation of the model deal better with competing costs, but it also offers more transparency behind the reasoning of the decisions being made in such circumstances. When we conceptualize our problem as a reinforcement learning task, an agent is tasked with making decisions in an environment with the aim of optimizing cumulative rewards (i.e., the total amount of reward it receives over the long run). Simply put, given discrete time steps $t = 0, 1, 2, 3, \dots$, at each time step the agent receives a representation of the environment's state, $s_t \in \altmathcal{S}$, and selects an action $a_t \in \altmathcal{A}(s_t)$, where $\altmathcal{A}(s_t)$ is the set of actions available in state $s_t$, and one step later receives a reward $r_{t+1} \in \altmathcal{R}$ and the state is updated~\cite{sutton2018reinforcement}. The way we define the way these rewards that are given to the agent, makes this decision process more transparent, however, it has its limitations. A universal optimal policy may not suit diverse socio-economic contexts due to variations in healthcare resources and economic vulnerabilities across countries, regions, or cities and a comprehensive consideration of decision factors, extending beyond pure reinforcement learning results is needed~\cite{Song2022,Dunn2017,Demir2006}.

Additionally, since most modern reinforcement learning achievements are due to a combination of deep learning~\cite{Mnih2015}, in the following framework we make the use of this. Deep reinforcement learning is an advancement to reinforcement learning which helps normalize the input and reduce its dimensionality~\cite{Lavet2018,Arulkumaran2017,Henderson_Islam_Bachman_Pineau_Precup_Meger_2018,Mnih2015}. We use a long short-term memory recurrent neural network for time-series data~\cite{NIPS2001_a38b1617,LSTM} and a simple fully connected network for data points that don't vary with time.

In summary, by using reinforcement learning augmented with deep learning techniques for the SIR with lockdown and time-varying vaccination rate environment, we can better understand the effects of lockdown measures on both public health outcomes and economic stability during a pandemic. However, it is crucial to consider the limitations of this approach and take into account a comprehensive set of decision factors in order to make informed policy decisions that are tailored to specific socio-economic contexts.

\section{Mathematical Formulation and Numerical Computation}
In this paper, we use a compartmental model to model infectious disease environment. We iteratively develop this model, starting with the foundational SIR model, to fit the actual data better. In an SIR model people of the population are divided based on whether they are yet to come into contact with an infected person (Susceptible), are infectious themselves (Infectious), or have recovered from the infection (Recovered). These compartments create the SIR model which can be represented as follows:
\subsection{Simple SIR Model}
\begin{figure}[htbp!]
\centering    
\begin{tikzpicture}[node distance=1cm, auto,
    >=Latex, 
    every node/.append style={align=center},
    int/.style={draw, minimum size=1cm}]
   \node [int] (S)             {$S$};
   \node [int, right=of S] (I) {$I$};
   \node [int, right=of I] (R) {$R$};
   \coordinate[right=of I] (out);
   \path[->] (S) edge node {$\lambda$} (I)
             (I) edge node {$\gamma$} (out);
\end{tikzpicture}
\end{figure}
\begin{equation}
  \frac{d S}{d t}=-\lambda S
  \label{eq:S_without_lockdown}
\end{equation}
\begin{equation}
  \frac{d I}{d t}=\lambda S-\gamma I
  \label{eq:I_without_lockdown}
\end{equation}
\begin{equation}
  \frac{d R}{d t}=\gamma I
  \label{eq:R_without_lockdown}
\end{equation}
Here, $\lambda$ is the force of infection, it is the rate at which susceptible individuals acquire an infectious disease~\cite{hens_aerts_faes_shkedy_lejeune_vandamme_beutels_2010}. It depends on other factors:
\begin{equation}
  \lambda = pc\frac{I}{N}
  \label{eq:lambda_force_of_infection}
\end{equation}
Here, $c$ is the average number of contacts a susceptible person makes per day. $p$ is the probability of the susceptible person becomes infectious after coming into contact with an infectious person. $\frac{I}{N}$ is the proportion of the contacts that are infectious.

And, $\beta$ the effective transmission rate is defined as:
\begin{equation}
  \beta = p c
  \label{eq:beta_effective_transmission_rate}
\end{equation}
During an epidemic, the fundamental drivers of an epidemic growth is the rate of infection $\beta$, i.e., the average number of infections per infected case and the infectious period $1/\gamma$, i.e., the average period for which the infected case is infected for. Epidemics can only happen if the case is infectious enough for long enough and this defined by $R_0 = \beta / \gamma$. Here, $R_0$ is the average number of secondary infections caused by each infected case, in an otherwise fully susceptible population.

At the peak of an epidemic, there is a decline as there are no more susceptible people left in the pool, therefore, $R_e$ (effective reproductive number) comes into play. $R_e$ is defined as the average number of secondary cases arising from an infected case, at a given point in an epidemic, therefore, it takes into account the existing immunity of the system~\cite{MASSAD2017232}. 
\begin{equation}
  R_e = R_0 \frac{S(t)}{N}
  \label{eq:R_effective}
\end{equation}
$S$ is the number of susceptible people and $N$ is the total population. At the start of an epidemic when everyone is susceptible, $R_e = R_0$ as, $S = N$ (i.e., the whole population is susceptible). $\beta$ and $\gamma$ are also used to define probability of and infectious individual infecting another individual $\beta / (\beta + \gamma)$ and the probability of recovery, $\gamma / (\beta + \gamma)$.

Most government policies look at the value of $R_e$ to come up with an effective strategy to combat the disease as the fate of the evolution of the disease depends upon it. When $R_e$ is less than one, the infected population $I$ will steadily decline to zero. Conversely, if $R_e$ is greater than one, the infected population will increase. In other words, when $\frac{dI(t)}{dt} < 0 \Rightarrow R_e < 1$ and $\frac{dI(t)}{dt} > 0 \Rightarrow R_e > 1$, therefore, the effective reproductive rate $R_e$ serves as a critical threshold that determines whether an infectious disease will rapidly extinguish or escalate into an epidemic~\cite{Cooper2020}.

To estimate the parameters $\beta$ and $\gamma$ for India based on the data from May 2020 to October 2022, we simply define two cost functions \cref{eq:cost_SIR,eq:cost_I} to calibrate the model with the use of Huber loss~\cite{huberloss}. Our model typically uses \cref{eq:cost_SIR}, considering all three compartments: susceptible, infected and recovered. However, there are instances where we need to balance modeling all groups and focusing on the infected group (population that drives disease spread). In such cases, we consider losses from both \cref{eq:cost_SIR,eq:cost_I}. A weighted sum of the loss functions allows for trade-offs between comprehensive modeling and focusing on infected group dynamics (see \cref{optimizing_window_length}, where a hyperparameter (window length) is selected keeping both these losses in mind).
\begin{equation}
  L_{\delta}(y, f(x)) = 
  \begin{cases}
    \frac{1}{2}{(y - f(x))^2} & \text{for } |y - f(x)| \leq \delta, \\
    \delta \cdot (|y - f(x)| - \frac{1}{2}\delta) & \text{otherwise.} \\
  \end{cases}
  \label{eq:huberloss}
\end{equation}
In the above equation, $y$ is the actual data and $f(x)$ is the prediction.
\begin{equation}
  \begin{split}
  loss\_SIR = \textrm{cost\_function\_SIR}(S, \hat{S}, I, \hat{I}, R, \hat{R}) \\
  = L_{\delta = 1}(S, \hat{S}) + L_{\delta = 1}(I, \hat{I}) + L_{\delta = 1}(R, \hat{R})
  \end{split}
  \label{eq:cost_SIR}
\end{equation}
\begin{equation}
  loss\_I = \textrm{cost\_function\_I}(I, \hat{I}) = L_{\delta = 1}(I, \hat{I})
  \label{eq:cost_I}
\end{equation}
Where, $S$ is the number of susceptible people and $\hat{S}$ is the predicted number of susceptible people, similarly, $I$ for infected and $\hat{I}$ for the predicted number of infected people, and $R$ for recovered. Using the equations \crefrange{eq:S_without_lockdown}{eq:R_without_lockdown}, we minimize the cost function \cref{eq:cost_SIR_without_lockdown} using the Nelder-Mead method~\cite{Gao2010} to estimate the parameters like $\beta$, $\gamma$ and fit the model to actual data. The following parameters and loss is obtained:
\begin{equation}
  \beta_{optimal} = 0.042
  \label{eq:beta_optimal_without_lockdown}
\end{equation}
\begin{equation}
  \gamma_{optimal} = 0.024
  \label{eq:gamma_optimal_without_lockdown}
\end{equation}
\begin{equation}
  R_0 = \frac{\beta_{optimal}}{\gamma_{optimal}} = 1.762
  \label{eq:r0_without_lockdown}
\end{equation}
\begin{equation}
  loss\_SIR = 85051490.533
  \label{eq:cost_SIR_without_lockdown}
\end{equation}
\begin{equation}
  loss\_I = 45187665.281
  \label{eq:cost_I_without_lockdown}
\end{equation}
See \cref{fig:SIR_model_IND_parent} to see how the model compares with the actual data. 

\subsection{SIR Model with Lockdown}
Now that a simple SIR model has been established -- we need to model the effects of the stringency index (measure for the strictness of lockdown) on $\beta$ (the effective transmission rate). To do this, we say the flow of susceptibles not only depend on $\beta$ but also $s(t)$ the stringency index at time~\cite{NBERw26867,AlvarezLockdownSIR,SIRLockdown,SIRLockdown2,analytical_covid_lockdown_model}. The stringency index is a composite measure based on nine response indicators including school closures, workplace closures, and travel bans, rescaled to a value from 0 to 100 (100 = strictest)~\cite{owidcoronavirus}. This index simply records the strictness of government policies and does not measure or imply the appropriateness or effectiveness of a country's response, i.e., a higher score does not necessarily mean that a country's response is "better" than others lower on the index.

To define the new time-varying beta that is dependent on the current stringency index, the following equations have been formulated:
\begin{equation}
  \frac{dS}{dt} = -\beta (1 - s(t)/100) \frac{S I}{N}
  \label{eq:S_with_lockdown}
\end{equation}
\begin{equation}
  \frac{dI}{dt} = \beta (1 - s(t)/100) \frac{S I}{N} - \gamma I
  \label{eq:I_with_lockdown}
\end{equation}
\begin{equation}
  \frac{dR}{dt} = \gamma I
  \label{eq:R_with_lockdown}
\end{equation}
Where, $s(t)$ is the stringency index at time $t$ and is scaled down by a factor of 100 to normalize it and bring it in the range $s(t)/100 \in [0, 1]$. Multiplying the rate of flow from $S$ to $I$ compartment with $1 - s(t)/100$ allows us to account for the effect that stringency has on the disease progression. A higher stringency index can theoretically stop the flow from the susceptible population to the infected population entirely. Optimizing these equations with \cref{eq:cost_SIR} using the Nelder-Mead method, we get the following parameters and loss:
\begin{equation}
  \beta_{optimal} = 0.401
  \label{eq:beta_optimal_with_lockdown}
\end{equation}
\begin{equation}
  \gamma_{optimal} = 0.090
  \label{eq:gamma_optimal_with_lockdown}
\end{equation}
\begin{equation}
  \begin{split}
    R_0 = \frac{\beta_{optimal}}{\gamma_{optimal}} (1 - s(t)) \\
    \overbar{R_0} = 1.693 \quad \text{(Mean)} \\
    \widetilde{R_0} = 1.624 \quad \text{(Median)} \\
    \text{Mode}(R_0) = 0.804 \quad \text{(Mode)} \\
    \sigma_{R_0} = 0.786 \quad \text{(Standard Deviation)} \\
    R_0 \in [0.16467, 3.0497] \quad \text{(Range)}
  \end{split}
  \label{eq:r0_with_lockdown}
\end{equation}
\begin{equation}
  loss\_SIR = 98438821.456
  \label{eq:cost_SIR_with_lockdown}
\end{equation}
\begin{equation}
  loss\_I = 11345389.686
  \label{eq:cost_I_with_lockdown}
\end{equation}
See \cref{fig:SIR_model_with_lockdown_IND_parent} to see how the model compares with the actual data. 

\subsection{SIR Model with Lockdown and Vaccination}
Lastly, an additional flow from the susceptible to recovered population can be shown by adding a vaccination rate $\nu$ in the model. 
\begin{equation}
  \frac{dS}{dt} = -\beta (1 - s(t)/100) \frac{S I}{N} - \nu S
  \label{eq:S_with_lockdown_and_nu}
\end{equation}
\begin{equation}
  \frac{dI}{dt} = \beta (1 - s(t)/100) \frac{S I}{N} - \gamma I
  \label{eq:I_with_lockdown_and_nu}
\end{equation}
\begin{equation}
  \frac{dR}{dt} = \gamma I + \nu S
  \label{eq:R_with_lockdown_and_nu}
\end{equation}
Optimizing these equations with \cref{eq:cost_SIR} using the Nelder-Mead method:
\begin{equation}
  \beta_{optimal} = 0.409
  \label{eq:beta_optimal_with_lockdown_and_nu}
\end{equation}
\begin{equation}
  \gamma_{optimal} = 0.092
  \label{eq:gamma_optimal_with_lockdown_and_nu}
\end{equation}
\begin{equation}
  \nu_{optimal} = 2.904 \times 10^{-5}
  \label{eq:nu_optimal_with_lockdown_and_nu}
\end{equation}
\begin{equation}
  \begin{split}
    R_0 = \frac{\beta_{optimal}}{\gamma_{optimal}} (1 - s(t)) \\
    \overbar{R_0} = 1.691 \\
    \widetilde{R_0} = 1.623 \\
    \text{Mode}(R_0) = 0.803 \\
    \sigma_{R_0} = 0.785 \\
    R_0 \in [0.165, 3.047]
  \end{split}
  \label{eq:r0_with_lockdown_and_nu}
\end{equation}
\begin{equation}
  loss\_SIR = 94636860.384
  \label{eq:cost_SIR_with_lockdown_and_nu}
\end{equation}
\begin{equation}
  loss\_I = 10840360.995
  \label{eq:cost_I_with_lockdown_and_nu}
\end{equation}

\subsection{Optimizing Window Length for Time-varying Vaccination Rate}\label{optimizing_window_length}
However, as observed by the value of $\nu_{optimal}$ from \cref{eq:nu_optimal_with_lockdown_and_nu}, which is almost negligible and the overestimation of infected individuals in \cref{fig:SIR_model_with_lockdown_with_vaccination_infections_IND} suggests that $\nu$ might be varying with time. This suggests to accurately estimate the infected population a time-varying vaccination rate should be used as the transition from susceptibility to direct recovery fluctuates with time~\cite{Liang2021,Marinov2022}. We first find the optimal window length~\cite{Liao2020} for which the value of $\nu$ is constant for a time sub-interval and results in the least loss from \cref{eq:cost_SIR,eq:cost_I}. For this we use different window lengths ($window\_lengths = 5, 10, 15 \dots 40, 45, 50 \textrm{ days}$).

\begin{equation}
  \frac{dS}{dt} = -\beta_{optimal}  (1 - s(t)/100)  \frac{S I}{N} - \nu S
  \label{eq:S_with_lockdown_and_nu_calc_nu}
\end{equation}
\begin{equation}
  \frac{dI}{dt} = \beta_{optimal}  (1 - s(t)/100) \frac{S I}{N} - \gamma_{optimal} I
  \label{eq:I_with_lockdown_and_nu_calc_nu}
\end{equation}
\begin{equation}
  \frac{dR}{dt} = \gamma_{optimal} I + \nu S
  \label{eq:R_with_lockdown_and_nu_calc_nu}
\end{equation}

\begin{figure}[htbp!]
  \begin{subfigure}[t]{0.48\textwidth}
    \centering
    \includegraphics[width=\linewidth]{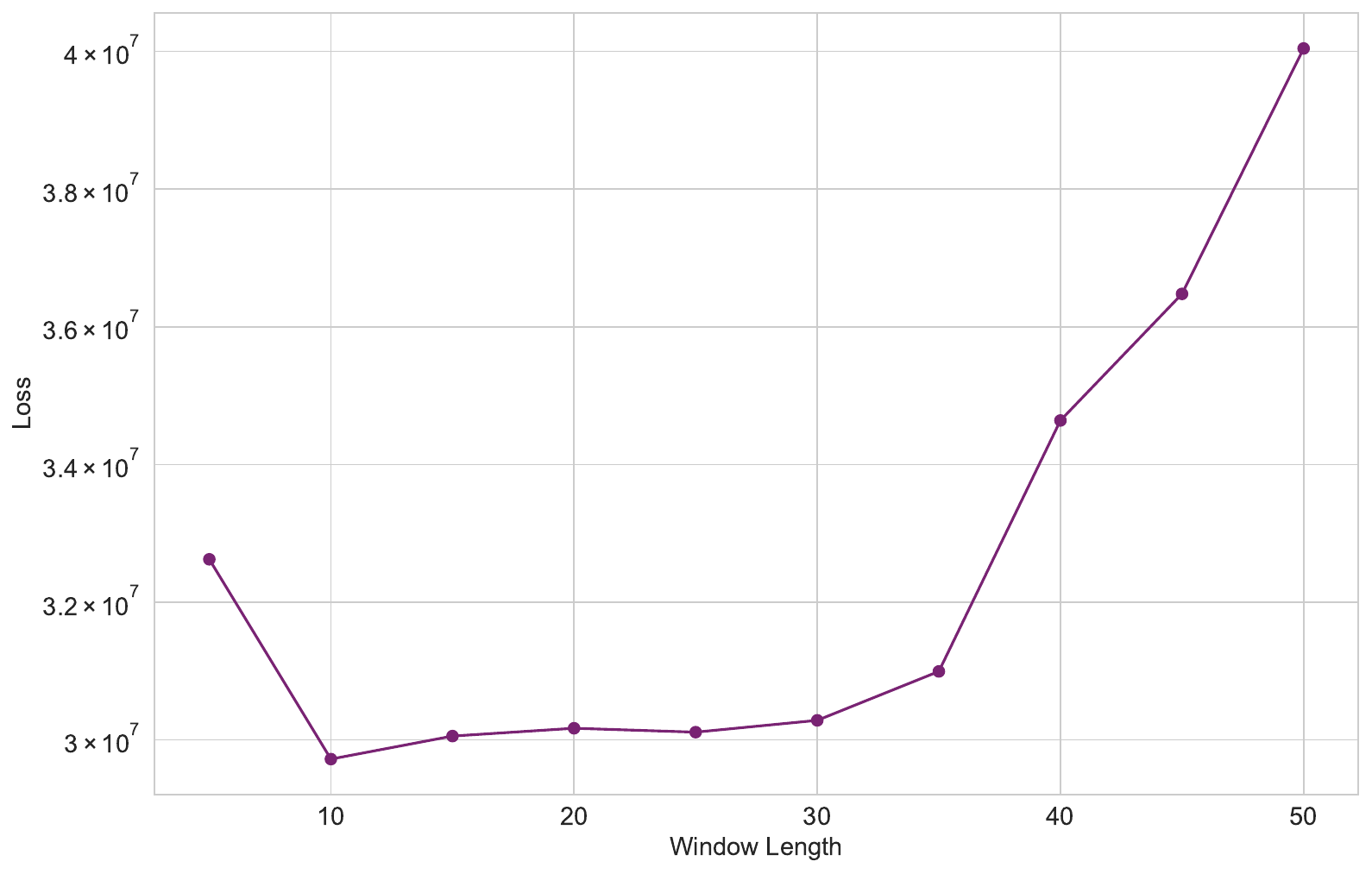}
    \caption{Loss for Different Window Lengths for Susceptible, Infected and Recovered Population}
    \label{fig:window_length_loss_SIR_IND}
  \end{subfigure}
  \hfill
  \begin{subfigure}[t]{0.48\textwidth}
    \centering
    \includegraphics[width=\linewidth]{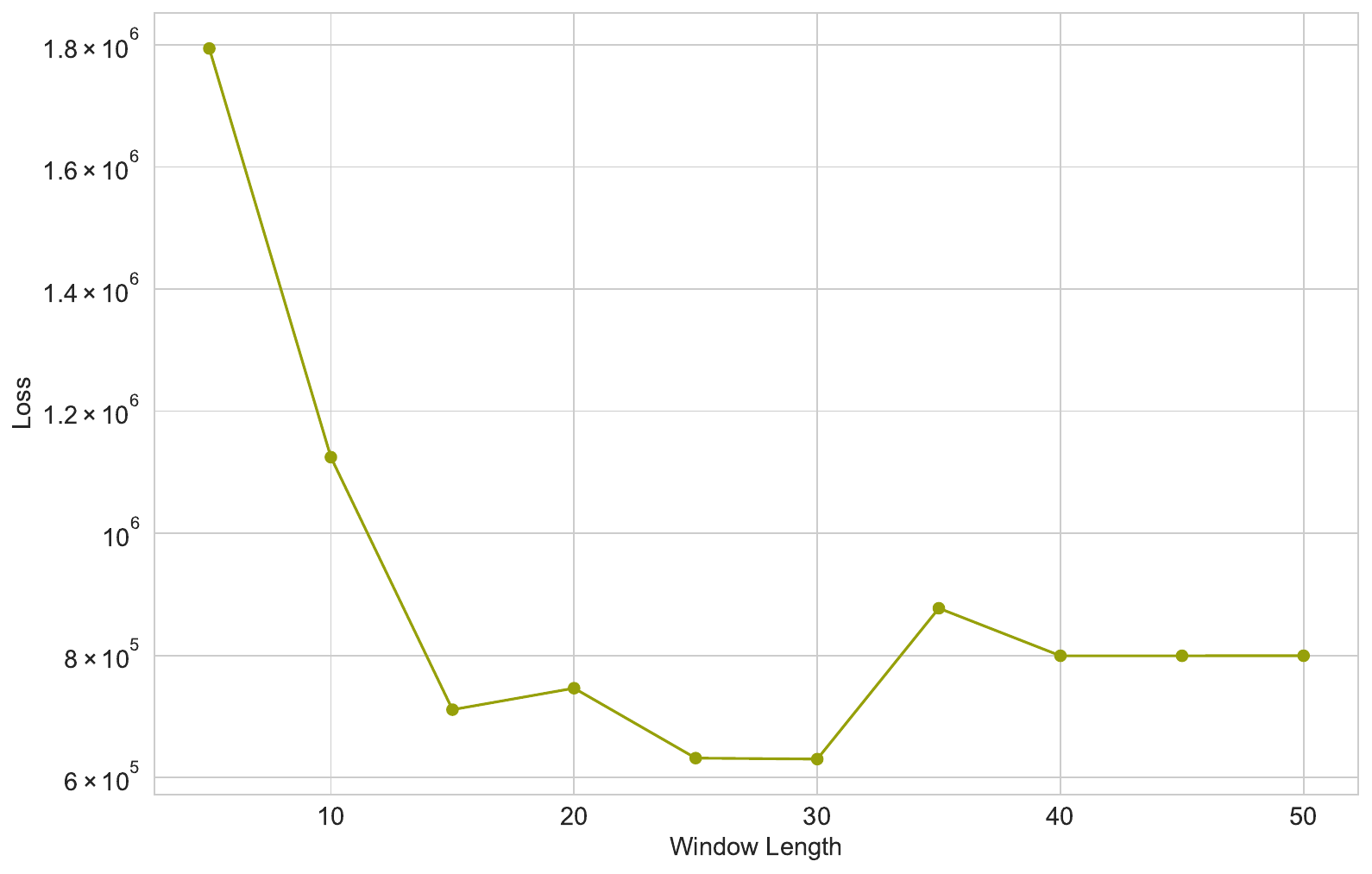}
    \caption{Loss for Different Window Lengths for Infected Population}
    \label{fig:window_length_loss_I_IND}
  \end{subfigure}
  \caption{\textbf{Loss for Different Window Lengths.} We try different window lengths to find the optimal loss for both cases, either when predicting all three populations (susceptibles, infected, recovered) or just the infected population.}
  \label{fig:window_length_loss_IND}
\end{figure}

Using the model described by \crefrange{eq:S_with_lockdown_and_nu_calc_nu}{eq:R_with_lockdown_and_nu_calc_nu} for each $window\_length_{i}$ in $window\_lengths$, where $i = 1, 2, 3 \dots 10$, we calculate the $time\_varying\_\nu$. For each $time\_varying\_\nu$ we then calculate the $loss\_SIR$ and $loss\_I$. This is done as follows:

\begin{enumerate}
  \item Let $start = 1$ (initial start day), $time\_varying\_\nu = [ ]$ (empty list).
  \item Repeat the following steps until $start + window\_length_{i}$ exceeds the total number of days in the data:
  \begin{enumerate}
    \item Estimate the value of $\nu$ for the sub-interval $[start, start + window\_length_{i}]$.
    \item Update $start = start + window\_length_{i}$.
    \item $time\_varying\_\nu.append(\nu)$
  \end{enumerate}
  \item Calculate $loss\_SIR$ and $loss\_I$ using the $time\_varying\_\nu$ using the $\beta_{optimal}$ and $\gamma_{optimal}$ from \cref{eq:beta_optimal_with_lockdown_and_nu} and \cref{eq:gamma_optimal_with_lockdown_and_nu}. The loss from the different $window\_lengths$ is compared in \cref{fig:window_length_loss_IND}.
\end{enumerate}

Note that the variable $\nu$ is constrained to be a positive integer, reflecting the inherent one-way nature of vaccination: individuals can only receive vaccinations, not return them.

The results in \cref{fig:window_length_loss_IND} indicate that a window length of $10$ days yields the least overall loss for all three population groups. However, this window length results in a poor approximation for just the infected group, which is crucial for accurately modelling the spread of the disease. Consequently, we have decided to use a window length of $15$ days, which provides a more accurate approximation for the infected population while still maintaining reasonable loss for the other groups.

\begin{figure}[htbp!]
  \centering
  \includegraphics[scale=0.35]{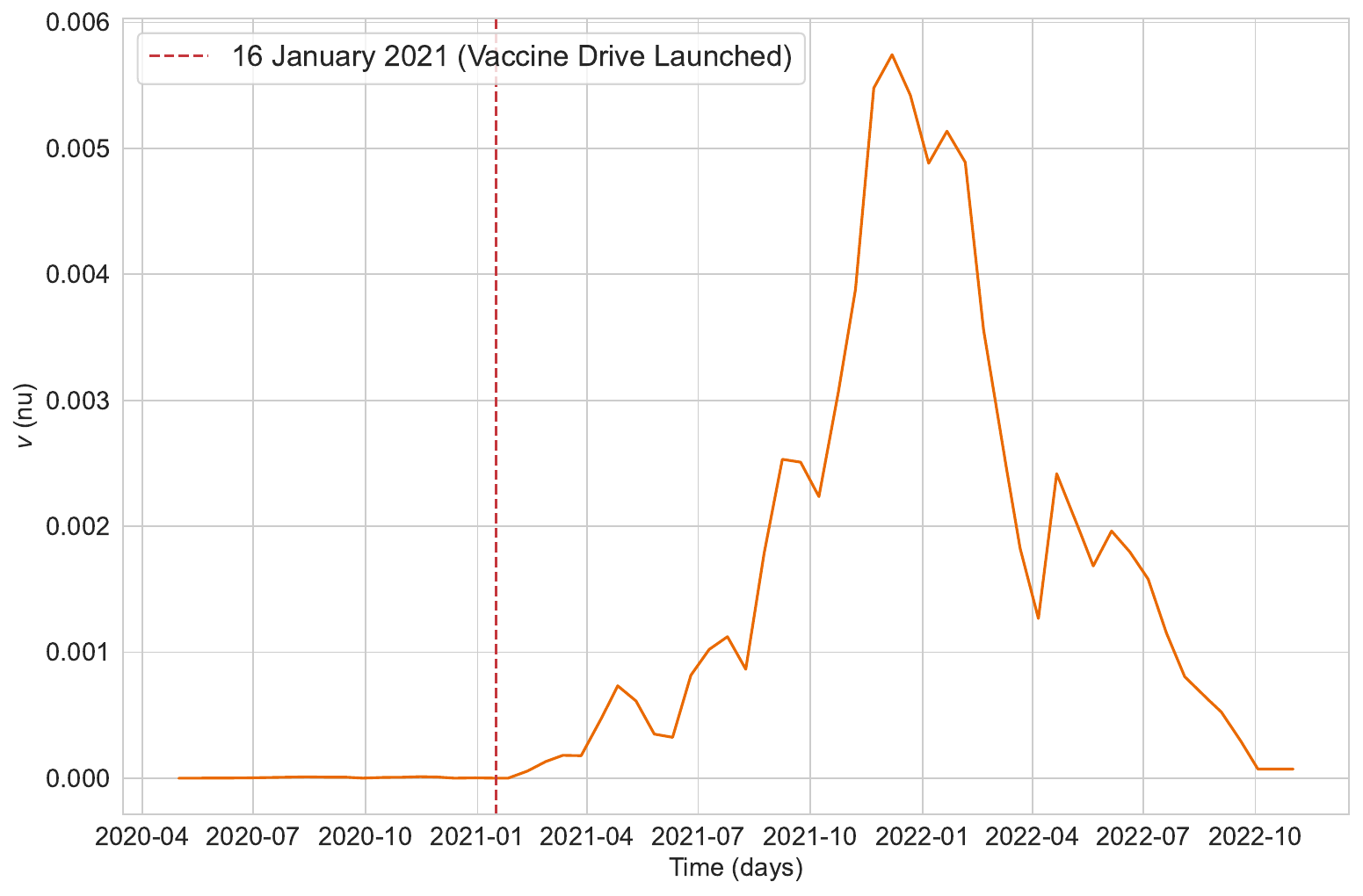}
  \caption{\textbf{$\nu$ Varying with Time.} This depicts how the vaccination rate ($\nu$) changes over time and highlights the introduction of the vaccination campaign in India.}
  \label{fig:interpolated_nu_varying_with_time_IND}
\end{figure}

\Cref{fig:interpolated_nu_varying_with_time_IND} shows us that the $\nu$ coincides with the actual of data of when the vaccination drive was first launched in India~\cite{UnicefCovidVaccine}. Therefore, using these values we finally recompute $\beta_{optimal}$ and $\gamma_{optimal}$ by supplying them into the equations for the SIR Model with lockdown and time-varying $\nu$.

See \cref{fig:SIR_model_with_lockdown_with_vaccination_IND_parent} to see how the model compares with the actual data.

\subsection{SIR Model with Lockdown and Time-varying Vaccination Rate}
\begin{figure}[htbp!]
	\begin{subfigure}[t]{0.48\textwidth}
		\centering
		\includegraphics[width=\linewidth]{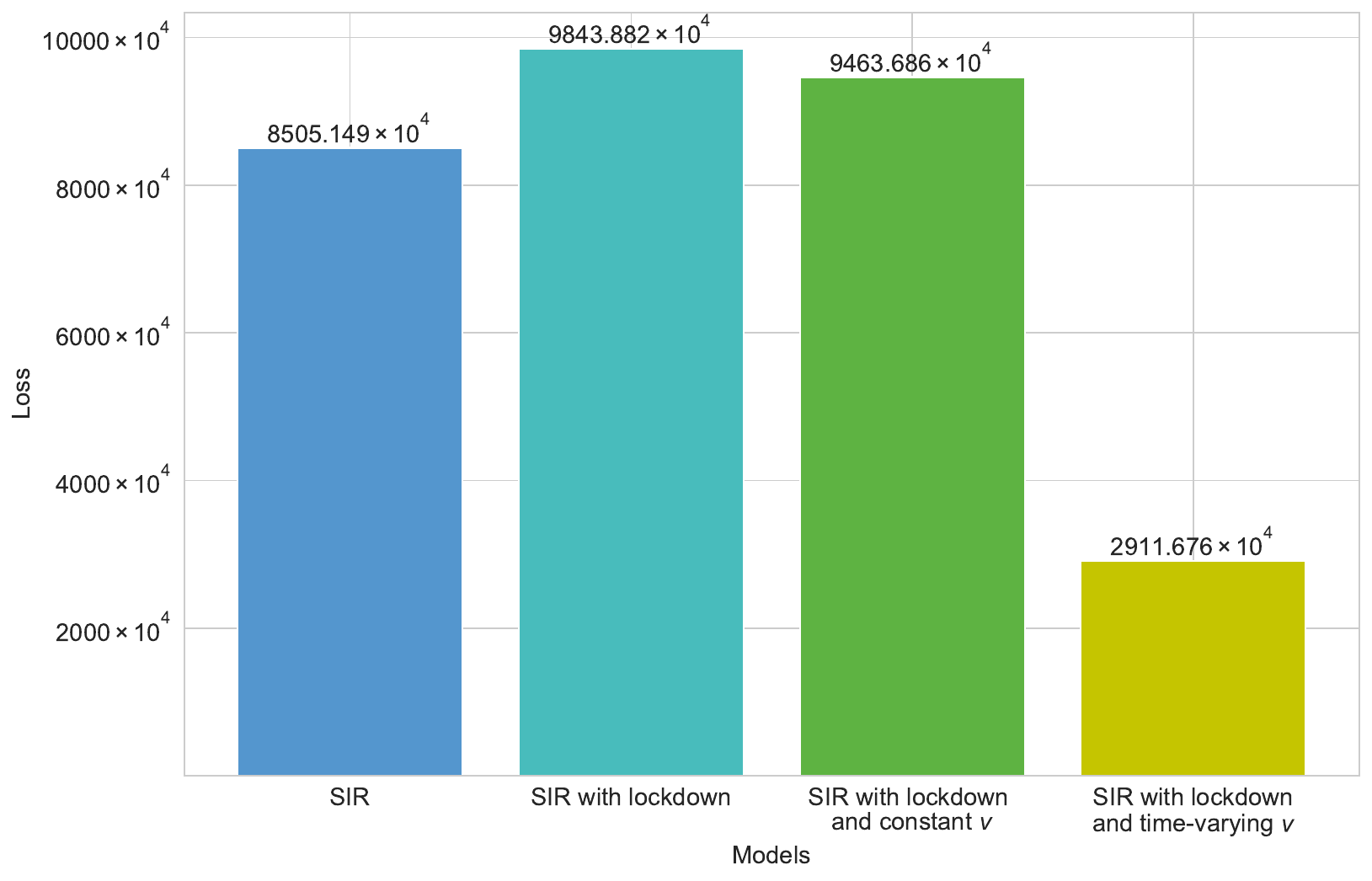}
		\caption{Loss for Different Models for Susceptible, Infected and Recovered Population}
		\label{fig:comparing_costs_SIR_IND}
	\end{subfigure}
	\hfill
	\begin{subfigure}[t]{0.48\textwidth}
		\centering
		\includegraphics[width=\linewidth]{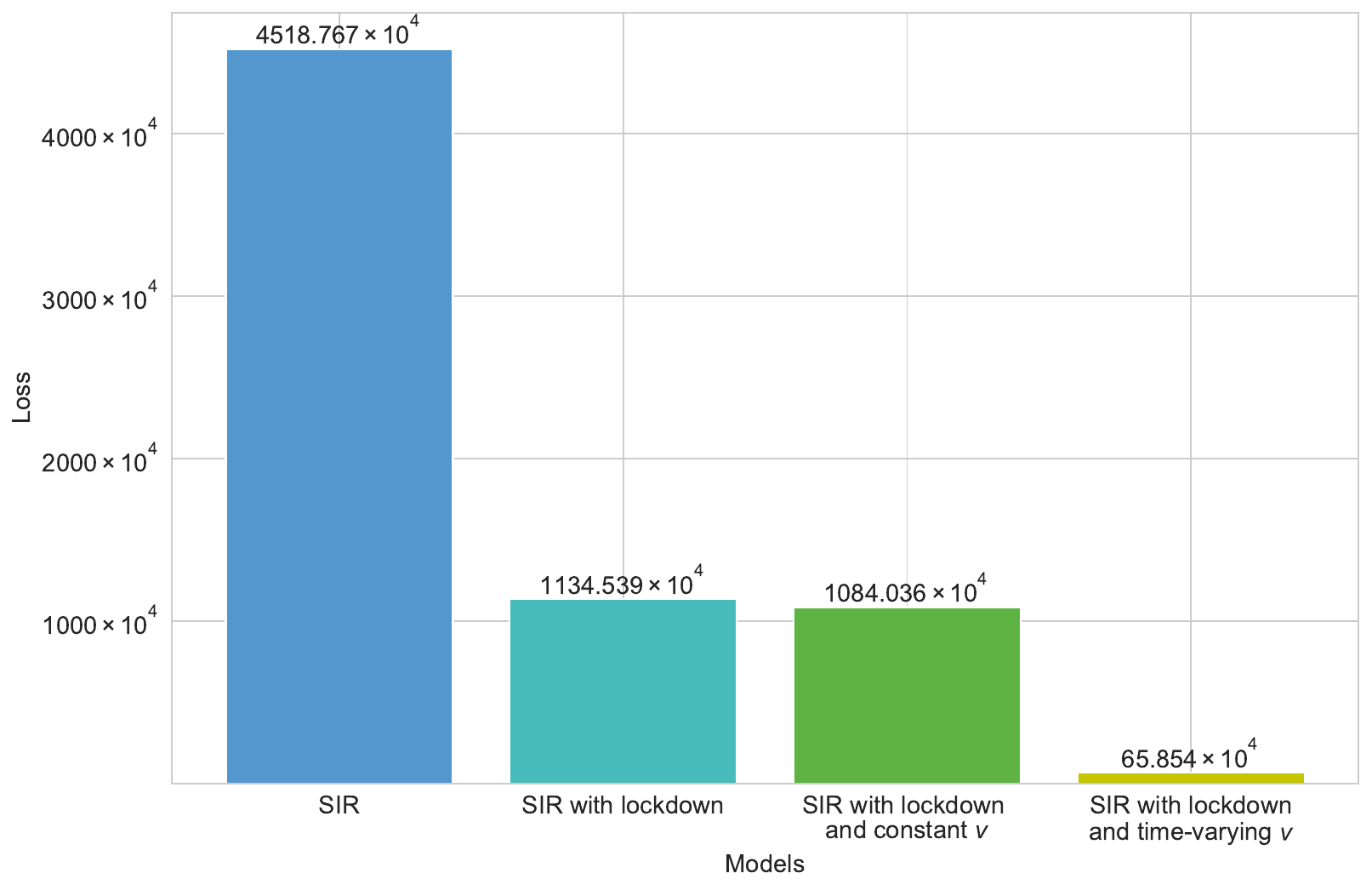}
		\caption{Loss for Different Models for Infected Population}
		\label{fig:comparing_costs_I_IND}
	\end{subfigure}
  \caption{\textbf{Loss for Different Models.} Here, we can observe that the loss is the least for the SIR model with lockdown and time-varying vaccination rate.}
  \label{fig:comparing_costs_IND_parent}
\end{figure}

Finally, we integrate the time-varying vaccination rate ($\nu$) into the SIR model that includes lockdown measures, resulting in the following set of equations, which represents our final model:
\begin{figure}[htbp!]
\begin{tikzpicture}[node distance=1cm, auto,
  >=Latex, 
  every node/.append style={align=center},
  int/.style={draw, minimum size=1cm}]
 \node [int] (S)             {$S$};
 \node [int, right=of S] (I) {$I$};
 \node [int, right=of I] (R) {$R$};
 \coordinate[right=of I] (out);
 \path[->] (S) edge node {$\lambda$} (I)
           (I) edge node {$\gamma$} (out)
           (S) edge[bend left=45] node[above] {$\nu(t)$} (R); 
\end{tikzpicture}
\end{figure}
\begin{equation}
  \frac{dS}{dt} = -\beta  (1 - s(t)/100)  \frac{S I}{N} - \nu(t) S
  \label{eq:S_with_lockdown_and_time_varying_nu}
\end{equation}
\begin{equation}
  \frac{dI}{dt} = \beta  (1 - s(t)/100) \frac{S I}{N} - \gamma I
  \label{eq:I_with_lockdown_and_time_varying_nu}
\end{equation}
\begin{equation}
  \frac{dR}{dt} = \gamma I + \nu(t) S
  \label{eq:R_with_lockdown_and_time_varying_nu}
\end{equation}
\begin{equation}
  \beta_{optimal} = 0.463
  \label{eq:beta_optimal_with_lockdown_and_time_varying_nu}
\end{equation}
\begin{equation}
  \gamma_{optimal} = 0.114
  \label{eq:gamma_optimal_with_lockdown_and_time_varying_nu}
\end{equation}
\begin{equation}
  \begin{split}
    \overbar{\nu_{optimal}} = 0.001 \\
    \widetilde{\nu_{optimal}} = 0.001 \\
    \text{Mode}(\nu_{optimal}) = 0.000 \\
    \sigma_{\nu_{optimal}} = 0.002 \\
    \nu_{optimal} \in [0.000, 0.006]
  \end{split}
  \label{eq:nu_optimal_with_lockdown_and_time_varying_nu}
\end{equation}
\begin{equation}
  \begin{split}
    R_0 = \frac{\beta_{optimal}}{\gamma_{optimal}} (1 - s(t)) \\
    \overbar{R_0} = 1.546 \\
    \widetilde{R_0} = 1.483 \\
    \text{Mode}(R_0) = 0.734 \\
    \sigma_{R_0} = 0.718 \\
    R_0 \in [0.150, 2.785]
  \end{split}
  \label{eq:r0_with_lockdown_and_time_varying_nu}
\end{equation}
\begin{equation}
  loss\_SIR = 29116762.926
  \label{eq:cost_SIR_with_lockdown_and_time_varying_nu}
\end{equation}
\begin{equation}
  loss\_I = 658537.443
  \label{eq:cost_I_with_lockdown_and_time_varying_nu}
\end{equation}
See \cref{fig:SIR_model_with_lockdown_with_vaccination_infections_time_varying_nu_IND_parent} to see how the model compares with the actual data and \cref{fig:comparing_costs_IND_parent} to see how the different models compare against each other.

\subsection{Modelling Normalized GDP with Stringency}
Now that a relation between $\beta$ and $s(t)$ is set up, it must be investigated how stringency index affects the normalized GDP~\cite{OECDNormalizedGDP,OECDNormalizedGDP2}. To do this, a polynomial equation of the third degree is fitted to the data points $f(x) = a x^3 + b x^2 + c x + d = y$, here, $x$ is the stringency ($s$) and $y$ the normalized GDP, and we minimize the squared error to find the values of coefficients $a, b, c, d$. For India after fitting a third-degree polynomial, the following equation is obtained:
\begin{equation}
    \begin{split}
      \textrm{normalized\_GDP} = -5.96640236 \times 10^{-5} s^{3} + 6.65064332 \times 10^{-3} s^{2} - 2.23109924 \times 10^{-1} s \\
    + 1.01357226 \times 10^{2}
    \end{split}
    \label{eq:gdp_modelled_with_stringency_IND}
\end{equation}

\subsection{Reinforcement Learning}
Given that the government is an agent that takes decisions in a deterministic environment defined above, we use reinforcement learning to model the competing costs of the environment. This environment is known as a Markov Decision Process (MDP) and is characterized by the Markov property. To possess the Markov property is to create a compact state signal that retains all relevant information from past sensations without requiring the complete history. The Markov property ensures that the probability of transitioning to the next state and receiving a reward depends only on the current state and action, without requiring the entire history~\cite{sutton2018reinforcement}. Our MDP is defined as follows:
\begin{itemize}
    \item Set of States $\altmathcal{S}$: The state of the environment is described through the descriptors like the normalized GDP, $R_e$, a list of all the previous actions (in changing the stringency) and the proportion of the population that was susceptible ($S$), infected ($I$) and recovered ($R$). The starting states are simply these values at the starting date and no previous actions.
    \item Actions $\altmathcal{A}$: The stringency index variable was analyzed with a sample size of 915. The mean value was approximately $61.96505$, with a standard deviation of $17.66983$. The minimum value was $31.48$, while the maximum value reached $96.3$. And the differences between two consecutive stringencies had a mean of $-0.070919$, and standard deviation of $1.42715$, with the minimum being $-14.36$ and maximum $16.67$. Based on this, we define the discrete action space. There are 7 actions for the agent, it can keep the stringency index same, reduce/increase by 2.5, reduce/increase by 5, and reduce/increase by 10 given that the stringency index doesn't exceed 100 or go below 0.
    \item Transition dynamics $\altmathcal{T}\left(\mathbf{s}_{t+1} \mid \mathbf{s}_t, \mathbf{a}_t\right)$ map a state-action pair at time $t$ onto a distribution of states at time $t+1$. This state transition is defined by the SIR model with lockdown and the model of how stringency index affects the GDP.
    \item Immediate reward $\altmathcal{R}\left(\mathbf{s}_t, \mathbf{a}_t, \mathbf{s}_{t+1}\right)$. The agent observes the state of the environment $\mathbf{s}_t$ at time $t$ and takes an action $\mathbf{a}_t$, after which the state transitions to $\mathbf{s}_{t+1}$ and the agent receives a reward $\mathbf{r}_{t+1}$ as feedback. In \cref{defining_reward_function} we define a reward strategy, however, it should be noted that this work serves as a framework where the strategy can be easily swapped for another to prioritize different needs.
    \item Discount Factor $\gamma \in[0,1]$, where lower values place more emphasis on immediate rewards. Here, we choose the default discount factor of 0.99.
\end{itemize}

Given that at each timestep $t$ the agent has to choose an action $a_t$ to maximize the reward $r_{t+1}$ a policy is formulized by the agent. The policy $\pi$ is a mapping from states to a probability distribution over actions: $\pi: \altmathcal{S} \rightarrow p(\altmathcal{A}=\mathbf{a} \mid \altmathcal{S})$. Reinforcement learning methods specify how the agent changes its policy as a result of its experience. If the MDP is episodic, i.e., the state is reset after each episode of length $T$, then the sequence of states, actions and rewards in an episode constitutes a trajectory or rollout of the policy. Every rollout of a policy accumulates rewards from the environment, resulting in the return $R=\sum_{t=0}^{t=T-1} \gamma^t r_{t+1}$. The goal of RL is to find an optimal policy, $\pi^*$, which achieves the maximum expected return from all states. To achieve this, reinforcement learning start with an initial arbitrary policy, i.e., a $Q$-table with no entries. $Q$-table is a mapping from states $s_t \in \mathscr{S}$ to a predefined set of actions to increase or decrease the stringency at time t, which are the actions $a_t \in \mathscr{A}$. Each entry of the $Q$-table $\left(Q_t\left(s_t, a_t\right)\right)$ associates an action in the finite sequence $\left(\mathscr{A}_j\right)_{j \in \mathbb{J}^{+}}$ to a state of the finite sequence $\left(\mathscr{S}_i\right)_{i \in \mathbb{I}^{+}}$~\cite{sutton2018reinforcement}. 

In this case of epidemic control by non-pharmaceutical interventions (NPI) based strategies this policy represents the series of stringencies to be imposed upon the population to shift the initial status of the environment to a targeted status which is equivalent to the desired set of system states. This is how the $Q$-table updates saying, if in state $s_k$ the most ideal action is $a_t$. After having more and more experience with the environment and understanding which actions lead to a higher reward $r$ an optimal policy is derived be maximizing the expected value of discounted reward $J\left(r_t\right)=\mathbb{E}\left[\sum_{t=1}^{\infty} \gamma^{(t-1)} r_t\right]$, where discount factor $\gamma \in [0, 1]$ (in our case $\gamma = 0.99$) and time steps $k = 1, 2, \dots$.

\subsubsection{Defining the Reward Function}\label{defining_reward_function}
The stringency index emerges as a critical factor influencing both the normalized GDP and the rate of infection spread. The decision to escalate or de-escalate the stringency index is a strategic one, with significant implications. Increasing the stringency decreases the spread of the infection. Conversely, it must be noted that herd immunity can only be achieved when the epidemic reaches its peak, i.e., when the effective reproductive number is equal to one ($R_e = 1$). This can only happen by lowering the stringency index which would allow the natural dynamics of the epidemic to transpire such that the population of susceptible individuals has depleted enough such that it is insufficient to propagate the disease further. Therefore, stringency is used to control the number of infected people and slow down the rate at which the epidemic reaches its peak, so that hospitals could house the number of infected people.

In reinforcement learning, positive rewards promote and negative rewards demote actions. The agent tries to generate such a policy/knowledge to avoid the discouraging situation by following the policy. By designing a proper reward function, it is possible to generate such an agent that may follow the human desired situation. While designing a reward function, it is important to note that the rewards we set up truly indicate what we want accomplished. In particular, the reward signal is not the place to impart to the agent prior knowledge about how to achieve what we want it to do~\cite{sutton2018reinforcement}. Taking inspiration from similar work~\cite{Ohi2020}, we define the reward function. 

The reward function is parameterized to account for key factors influencing decision-making. To incentivize reduction of $R_e$ (effective reproductive number) and the increase of the normalized GDP after $R_e$ is below $1.5$. The reward is defined as follows:

\[
\textrm{Reward} = \begin{cases}
-20 \times R_e & \textrm{if } R_e > 1.5 \\
100 \times \textrm{min\_max\_normalized\_GDP} & \textrm{if } 1.25 \leq R_e \leq 1.5 \\
200 \times \textrm{min\_max\_normalized\_GDP} & \textrm{if } R_e < 1.25
\end{cases}
\]

This reward function is parameterized to account for key factors influencing decision-making. When the effective reproductive number ($R_e$) exceeds $1.5$, indicating a high transmission rate of the disease, the reward is negatively impacted to incentivize a reduction in $R_e$. As $R_e$ decreases within the range $1.25 \leq R_e \leq 1.5$, indicating a moderate transmission rate, the reward is directly proportional to the normalized GDP, reflecting the importance of both controlling the spread of the disease and maintaining economic stability. Notably, when $R_e$ drops below $1$, signaling a declining transmission rate and potential containment of the disease, the reward function shifts focus towards economic recovery. In this scenario, the reward incentivizes an increase in the normalized GDP, emphasizing the need to stimulate economic activity and promote recovery efforts following successful control measures.

Additionally, if the proportion of the infected population were to rise above 0.003 (peak in the actual data) the model is punished ($-2000$) and otherwise rewarded ($50$). To reward not changing the stringencies frequently, we reward the absolute different between the previous stringency and the current stringency negatively ($|s(t) - s(t-1)| \times -12$).

It should be realized there can be an infinite number of ways to design the reward function to be more human and upgrade the way a decision is taken given the situation~\cite{AWSDeepRacer}. Therefore, this research acts as a  framework for promoting the development of more efficient reward strategies for the same. 

\subsubsection{Deep Reinforcement Learning and Training}\label{drl_and_training}
\begin{figure}[htbp!]
  \centering
  \includegraphics[width=0.7\linewidth]{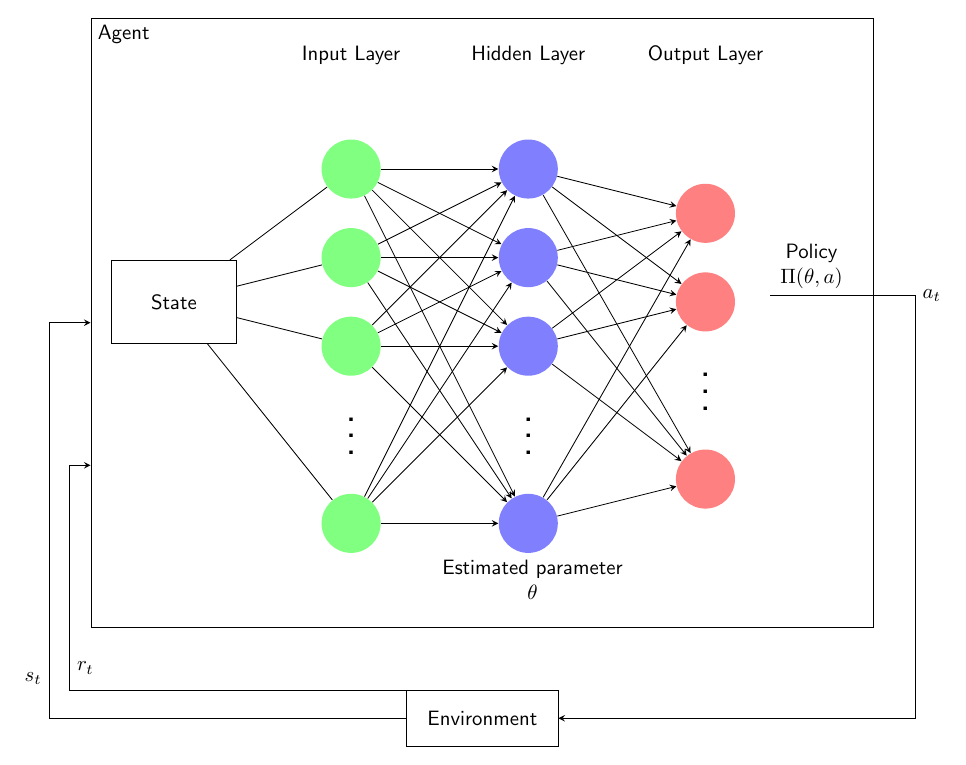}
  \caption{\textbf{Deep Reinforcement Learning.} Deep learning algorithms used in reinforcement learning enables more complex decision-making.}
  \label{fig:sir_arch}
\end{figure}

The agent observes the percentage of the population that is susceptible, infected, recovered, and time-varying data like the GDP, and previous actions taken to change the stringency, and $R_e$. Since Stable Baselines3 can support multiple inputs (time-series data, single data points and images) by using Dict Gym space. For data that varies with time (stringency, normalized GDP, $R_e$) we use a simple long short-term memory architecture~\cite{LSTM}. For other data, such as the current proportion of the population that is susceptible, infected, and recovered, we use a simple fully connected layer. The output from both of these networks is concatenated and used by the reinforcement learning agent for training. The schematic diagram of neural networks used to inform reinforcement learning agent's decision-making process is given in \cref{fig:sir_arch}. We train the model for $2742$ time steps and some of the best results are presented.

\section{Results}
\begin{figure}[htbp!]
  \centering
  \begin{subfigure}[t]{\textwidth}
    \centering
    \includegraphics[scale=0.35]{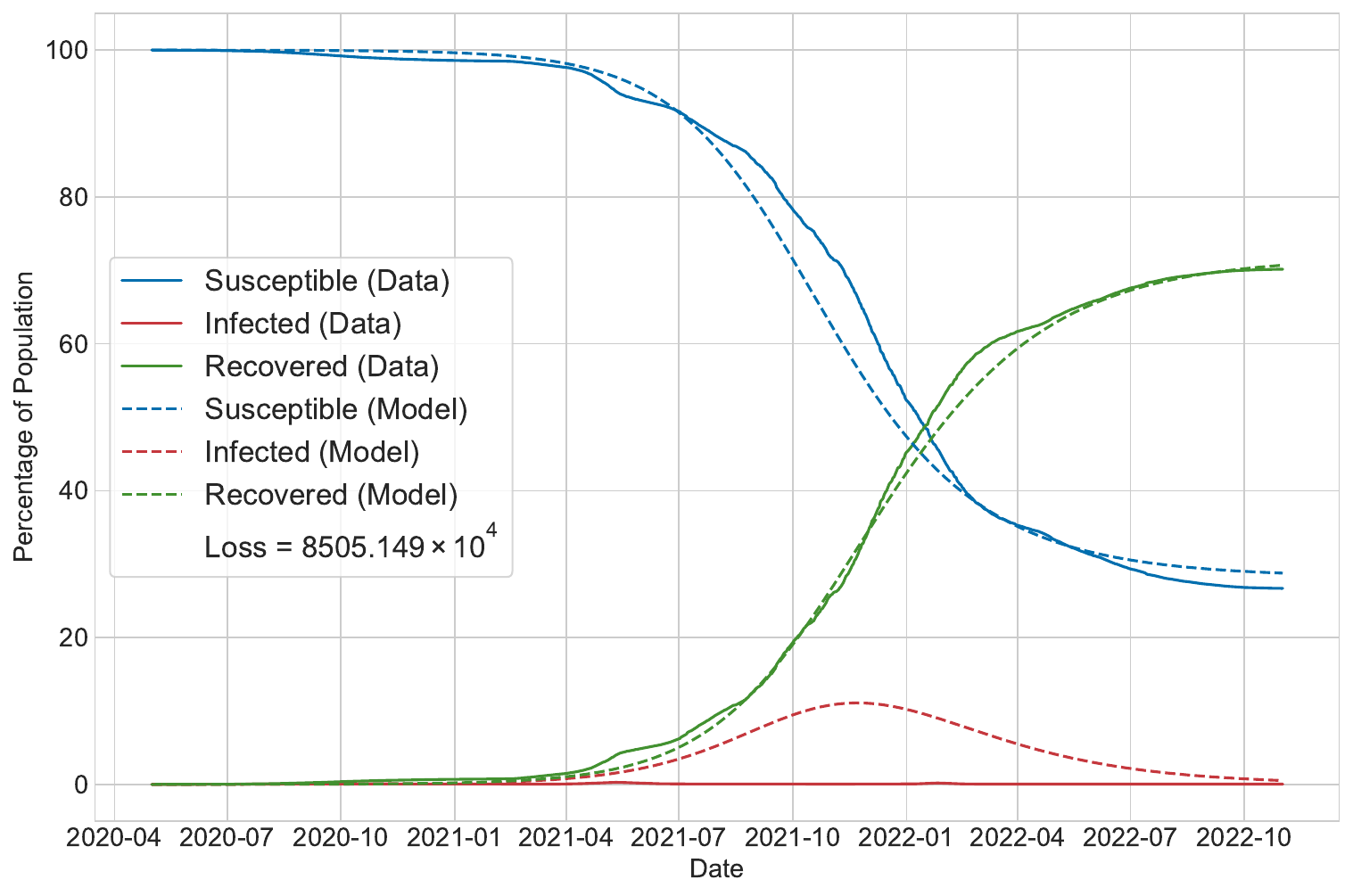}
    \caption{SIR Model}
    \label{fig:SIR_model_IND}
  \end{subfigure}
  \begin{subfigure}[t]{\textwidth}
    \centering
    \includegraphics[scale=0.35]{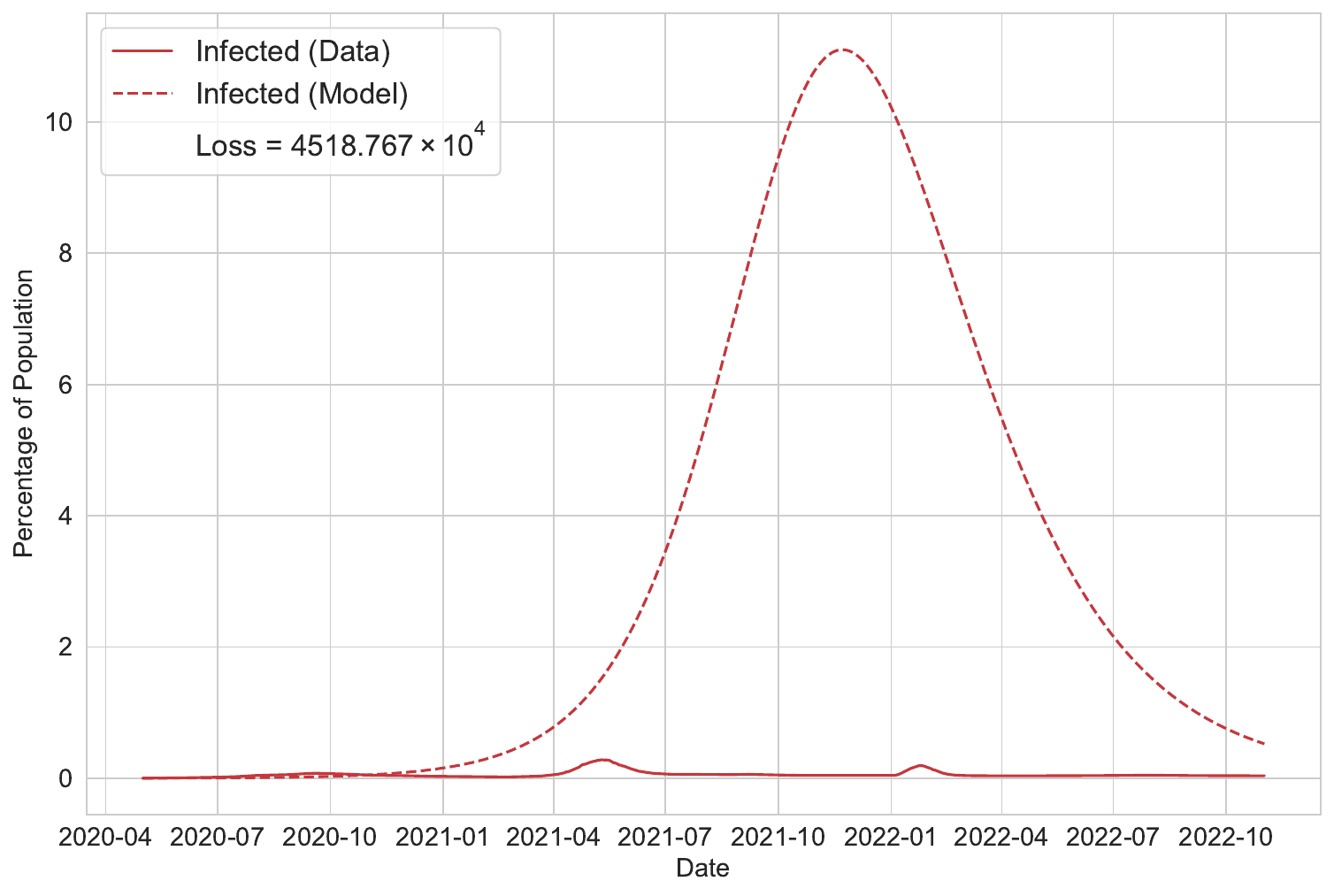}
    \caption{Infections Modelled with SIR Model}
    \label{fig:SIR_model_infections_IND}
  \end{subfigure}
  \caption{\textbf{SIR Model Comparison for India.} The figure presents a comparison between the fitted simple SIR model (\crefrange{eq:S_without_lockdown}{eq:cost_I_without_lockdown}) and real data. Here, an evident overestimation of the infected population is observed.}
  \label{fig:SIR_model_IND_parent}
\end{figure}

Using the simple SIR model from \crefrange{eq:S_without_lockdown}{eq:cost_I_without_lockdown},  to model the disease dynamics we get \cref{fig:SIR_model_IND_parent}. Here, it can be observed that the SIR model accurately fits the susceptible population and recovered population but overestimates the infected population by a significant margin which can create complications. This is because disease dynamics are controlled by this population and our work involves rewarding the agent when the proportion of infected individuals falls below a predetermined threshold. Therefore, an overestimation of the infected population could lead to incorrect decision-making and undesirable outcomes.

\begin{figure}[htbp!]
  \centering
  \begin{subfigure}[t]{\textwidth}
    \centering
    \includegraphics[scale=0.35]{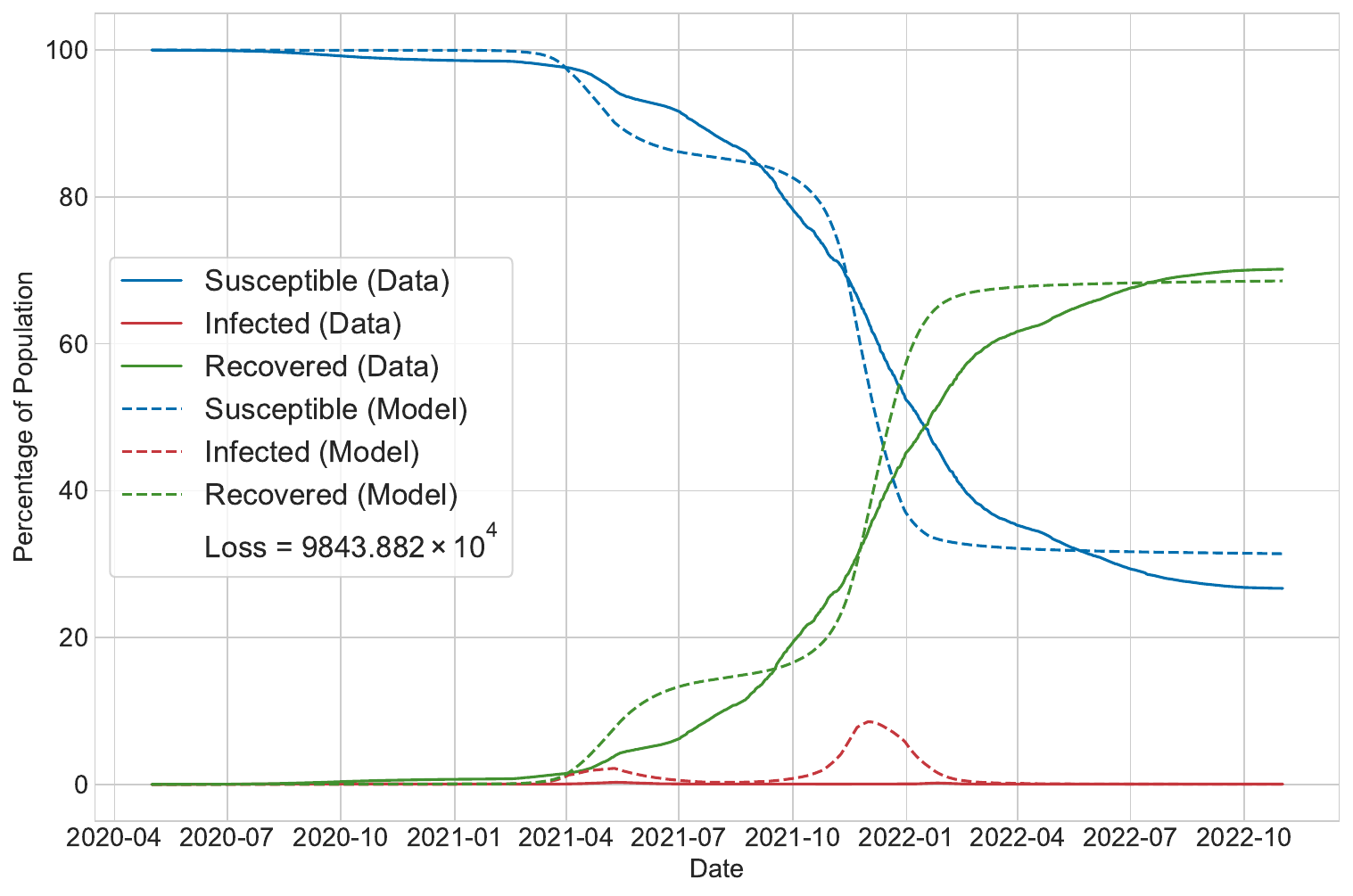}
    \caption{SIR Model with Lockdown}
    \label{fig:SIR_model_with_lockdown_IND}
  \end{subfigure}
  \begin{subfigure}[t]{\textwidth}
    \centering
    \includegraphics[scale=0.35]{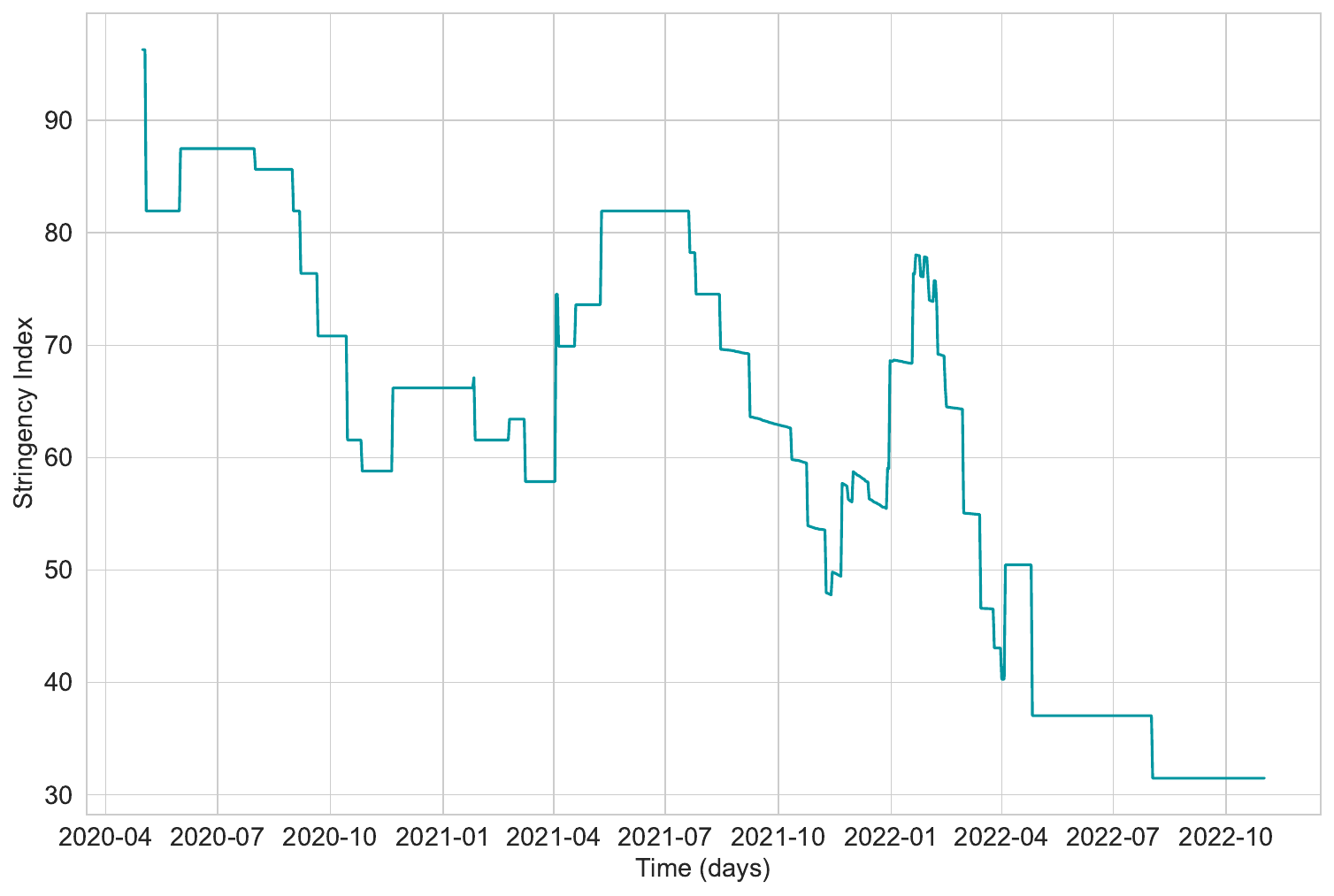}
    \caption{Stringency Varying with Time}
    \label{fig:stringency_varying_with_time_IND}
  \end{subfigure}
  \begin{subfigure}[t]{\textwidth}
    \centering
    \includegraphics[scale=0.35]{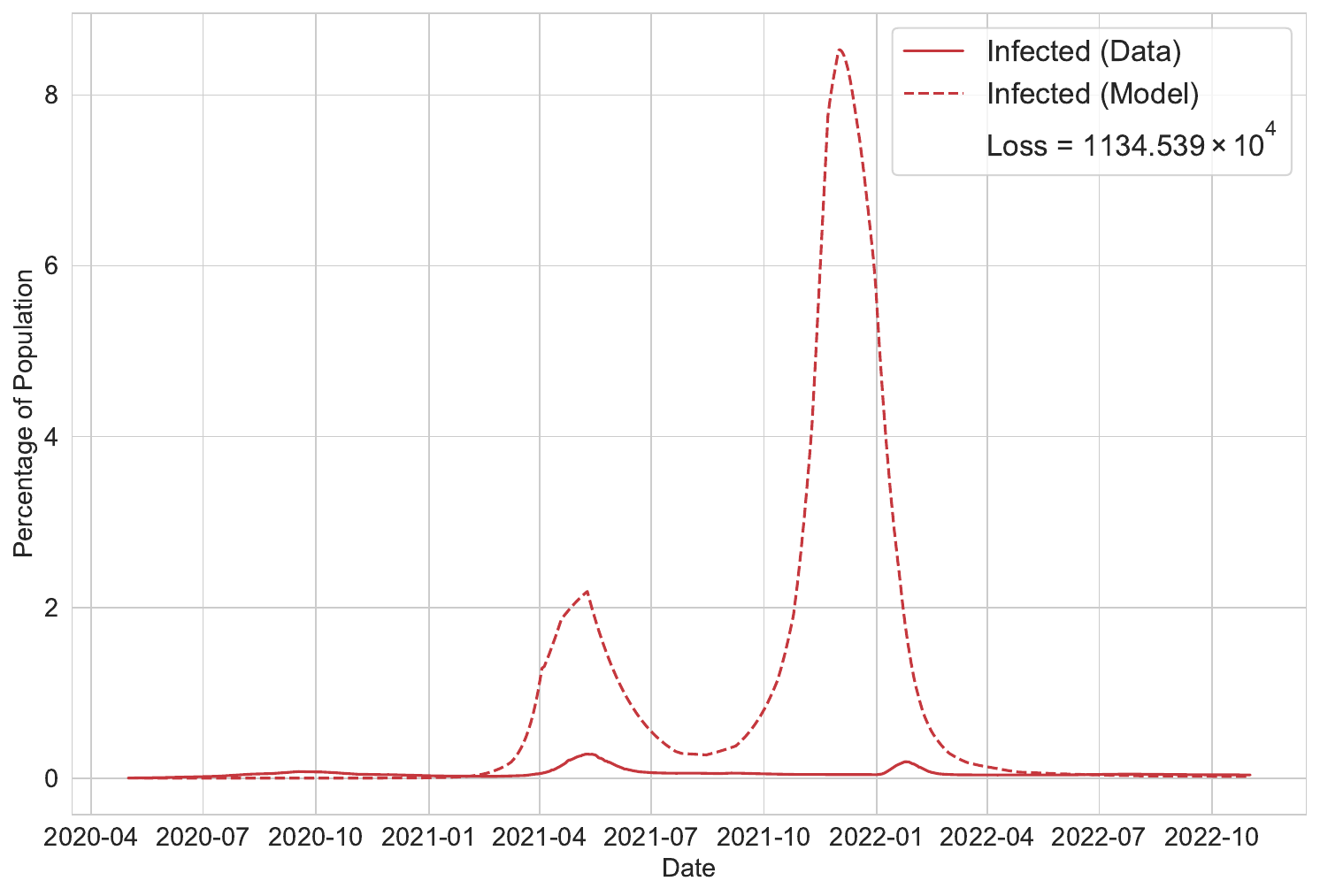}
    \caption{Infections Modelled with SIR Model with Lockdown}
    \label{fig:SIR_model_with_lockdown_infections_IND}
  \end{subfigure}
  \caption{\textbf{SIR Model with Lockdown Analysis for India.} This figure illustrates the fitting of the SIR model with lockdown (\crefrange{eq:S_with_lockdown}{eq:cost_I_with_lockdown}) in comparison to real data. The introduction of lockdown measures showcases discernible effects on the dynamics of disease progression. While an overestimation persists, the model's peaks now closely align with the observed data and is able to capture key trends.}
  \label{fig:SIR_model_with_lockdown_IND_parent}
\end{figure}

Combining the lockdown dynamics in the SIR model using \crefrange{eq:S_with_lockdown}{eq:cost_I_with_lockdown}, we get the following \cref{fig:SIR_model_with_lockdown_IND_parent}. Here, it can be observed there's an overestimation of infected individuals, but, the two stages of the epidemic are being accounted for. This is what suggests that there might be depletion of infected individuals through vaccination.

\begin{figure}[htbp!]
  \centering
  \begin{subfigure}[t]{\textwidth}
    \centering
    \includegraphics[scale=0.35]{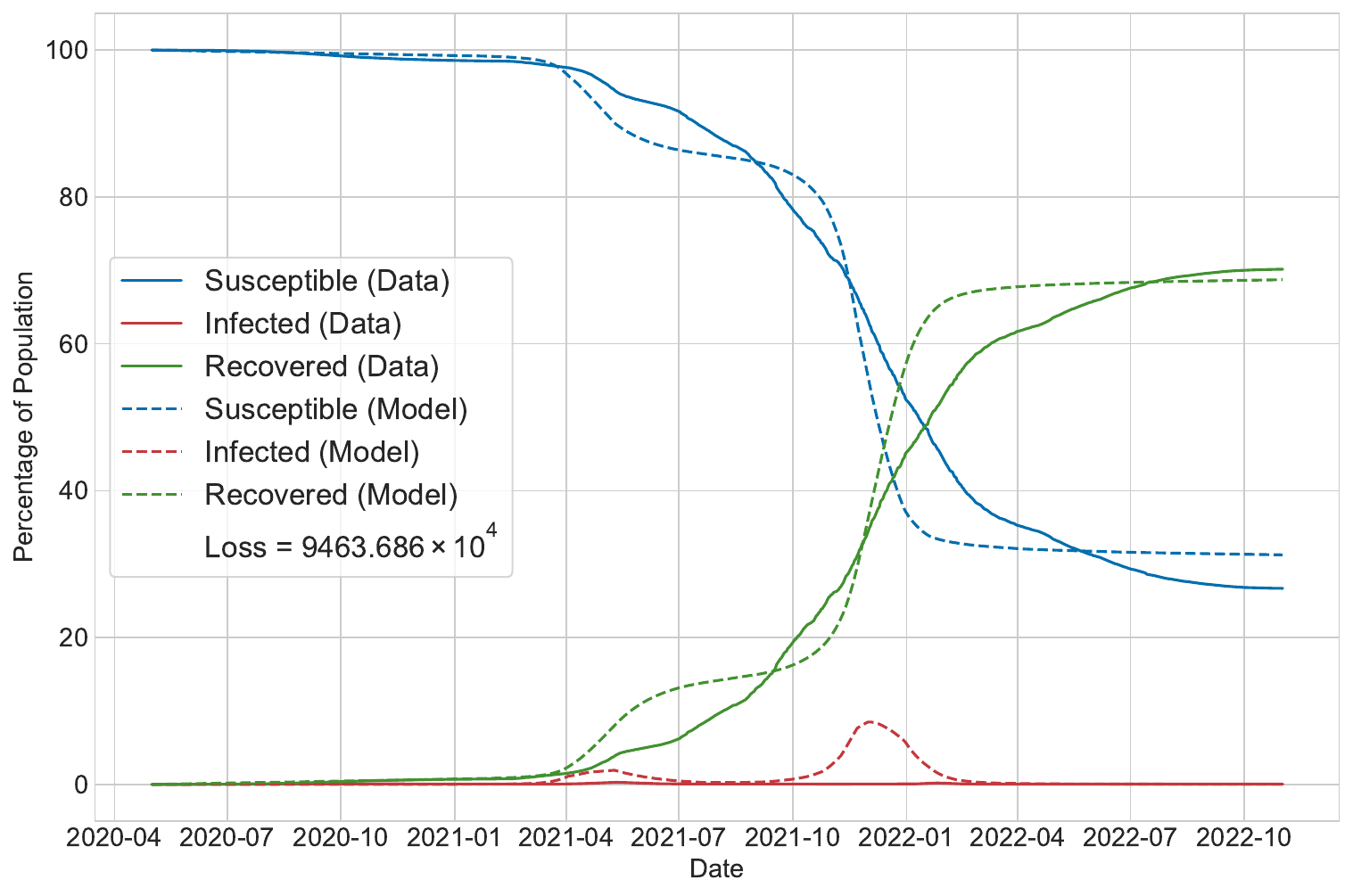}
    \caption{SIR Model with Lockdown and Vaccination}
    \label{fig:SIR_model_with_lockdown_with_vaccination_IND}
  \end{subfigure}
  \begin{subfigure}[t]{\textwidth}
    \centering
    \includegraphics[scale=0.35]{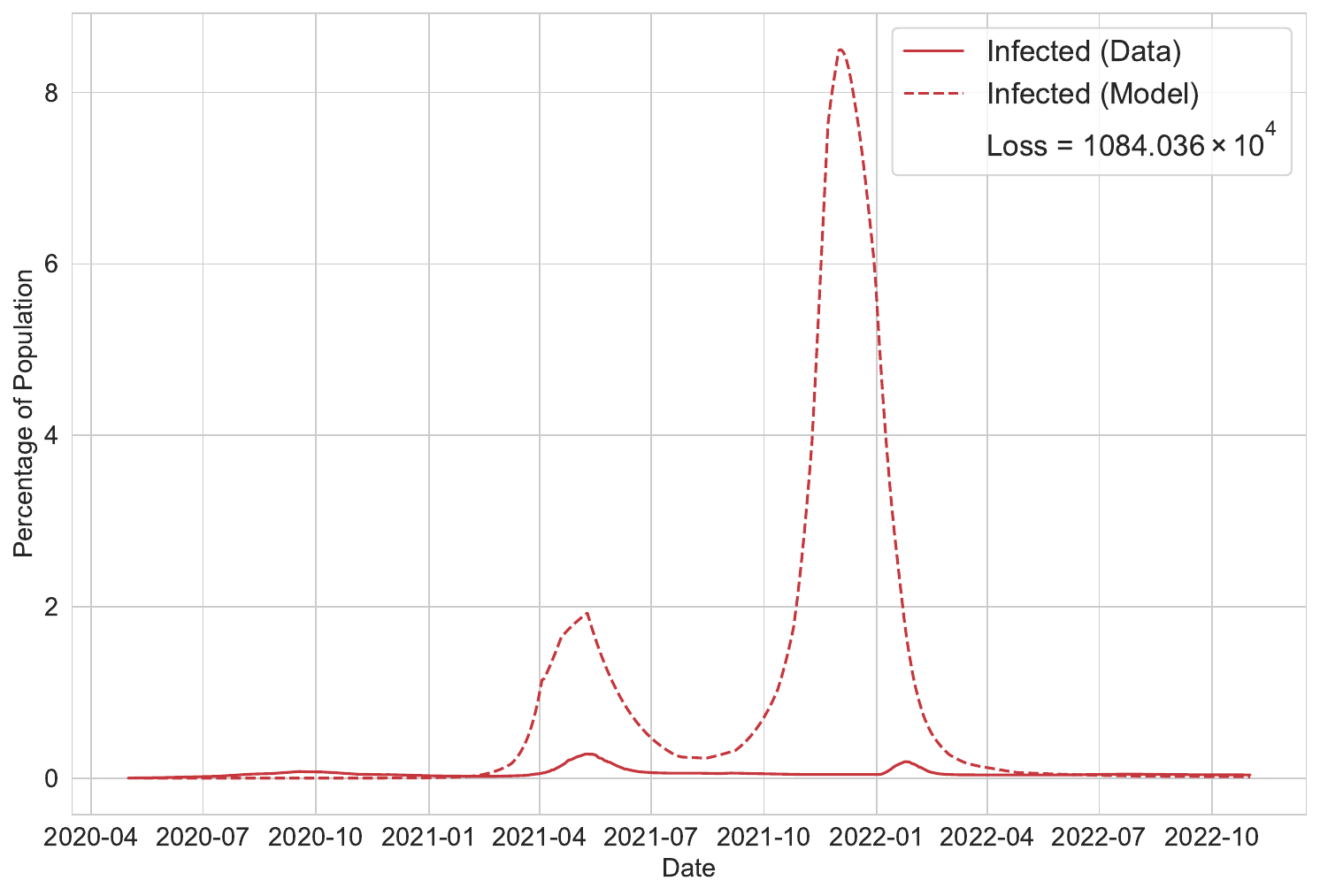}
    \caption{Infections Modelled with SIR Model with Lockdown and Vaccination}
    \label{fig:SIR_model_with_lockdown_with_vaccination_infections_IND}
  \end{subfigure}
  \caption{\textbf{SIR Model with Lockdown and Vaccination for India.} This figure displays the fitting of the SIR model with lockdown (\crefrange{eq:S_with_lockdown_and_nu}{eq:cost_I_with_lockdown_and_nu}) compared to the real data. The infection trends closely resemble those depicted by the SIR model with lockdown, as illustrated in \cref{fig:SIR_model_with_lockdown_infections_IND} and this is because the rate of vaccination is negligible \cref{eq:nu_optimal_with_lockdown_and_nu}. This is suggestive of a rate of vaccination that varies with time.}
  \label{fig:SIR_model_with_lockdown_with_vaccination_IND_parent}
\end{figure}

Incorporating vaccination dynamics into the SIR model with lockdown measures, as described by \crefrange{eq:S_with_lockdown_and_nu}{eq:cost_I_with_lockdown_and_nu}, we get the following \cref{fig:SIR_model_with_lockdown_with_vaccination_IND_parent}. Here, because the value of $\nu$ \cref{eq:nu_optimal_with_lockdown_and_nu} is negligible, it doesn't change the results significantly compared to the previous model (\crefrange{eq:S_with_lockdown}{eq:cost_I_with_lockdown} and \cref{fig:SIR_model_with_lockdown_IND_parent}). Therefore, a time-varying $\nu$ shall be able to better account for the these dynamics.

\begin{figure}[htbp!]
  \centering
  \begin{subfigure}[t]{\textwidth}
    \centering
    \includegraphics[scale=0.35]{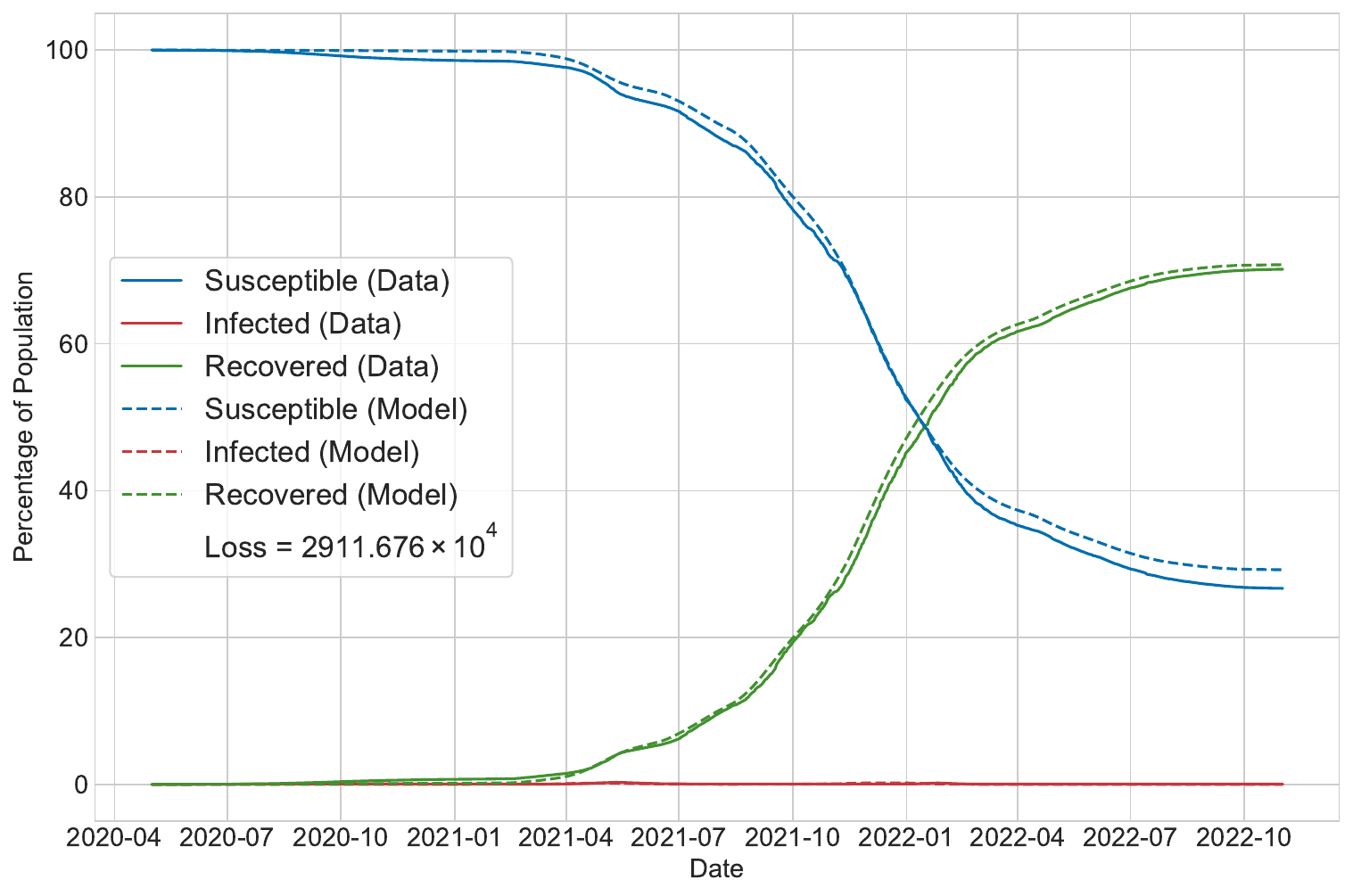}
    \caption{SIR Model with Lockdown and Time-varying Vaccination Rate}
    \label{fig:SIR_model_with_lockdown_with_vaccination_time_varying_nu_IND}
  \end{subfigure}
  \begin{subfigure}[t]{\textwidth}
    \centering
    \includegraphics[scale=0.35]{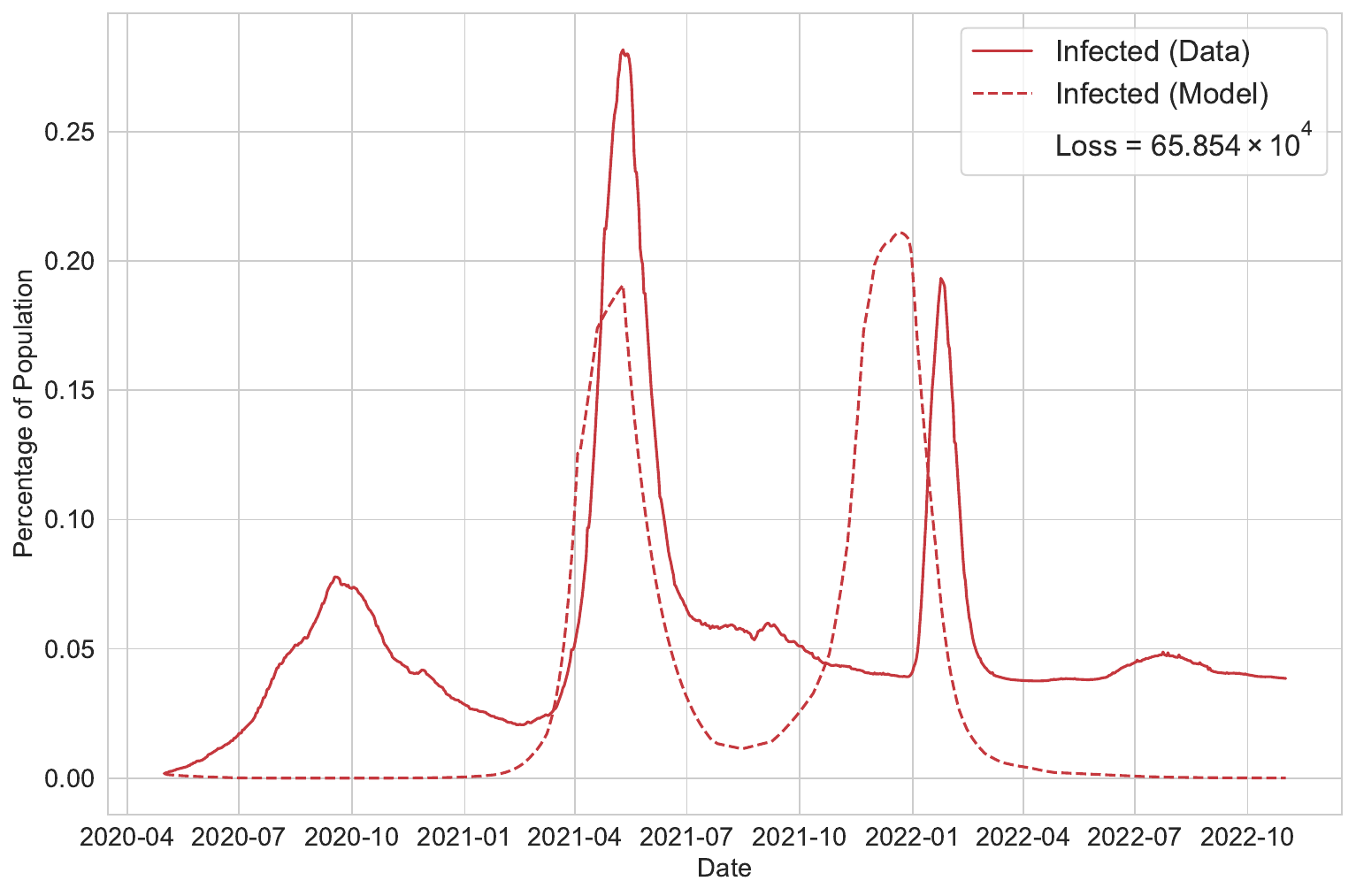}
    \caption{Infections Modelled with SIR Model with Lockdown and Time-varying Vaccination Rate}
    \label{fig:SIR_model_with_lockdown_with_vaccination_infections_time_varying_nu_IND}
  \end{subfigure}
  \caption{\textbf{SIR Model with Lockdown and Time-varying Vaccination Rate.} This figure displays the fitting of the SIR model with lockdown and time-varying vaccination rate (\crefrange{eq:S_with_lockdown_and_time_varying_nu}{eq:cost_I_with_lockdown_and_time_varying_nu}) compared to the real data.  Incorporating a time-varying vaccination rate enhances the model's ability to capture variations in the infected population over time.}
  \label{fig:SIR_model_with_lockdown_with_vaccination_infections_time_varying_nu_IND_parent}
\end{figure}

For SIR model with lockdown and time-varying vaccination rate from \crefrange{eq:S_with_lockdown_and_time_varying_nu}{eq:cost_I_with_lockdown_and_time_varying_nu}, we get the following \cref{fig:SIR_model_with_lockdown_with_vaccination_infections_time_varying_nu_IND_parent}. With a time-varying $\nu$ (vaccination rate) and the effect of lockdown, our model is able to account for the infected individuals and reduce the cost in comparison to all the previously formalized models for the data. This shows how interventions and changes in the way people behave in response of an epidemic~\cite{Caldwell2021} play a major role in the way the epidemic unfolds.

\begin{figure}[htbp!]
  \centering
  \begin{subfigure}[t]{0.48\textwidth}
    \centering
    \includegraphics[width=\linewidth]{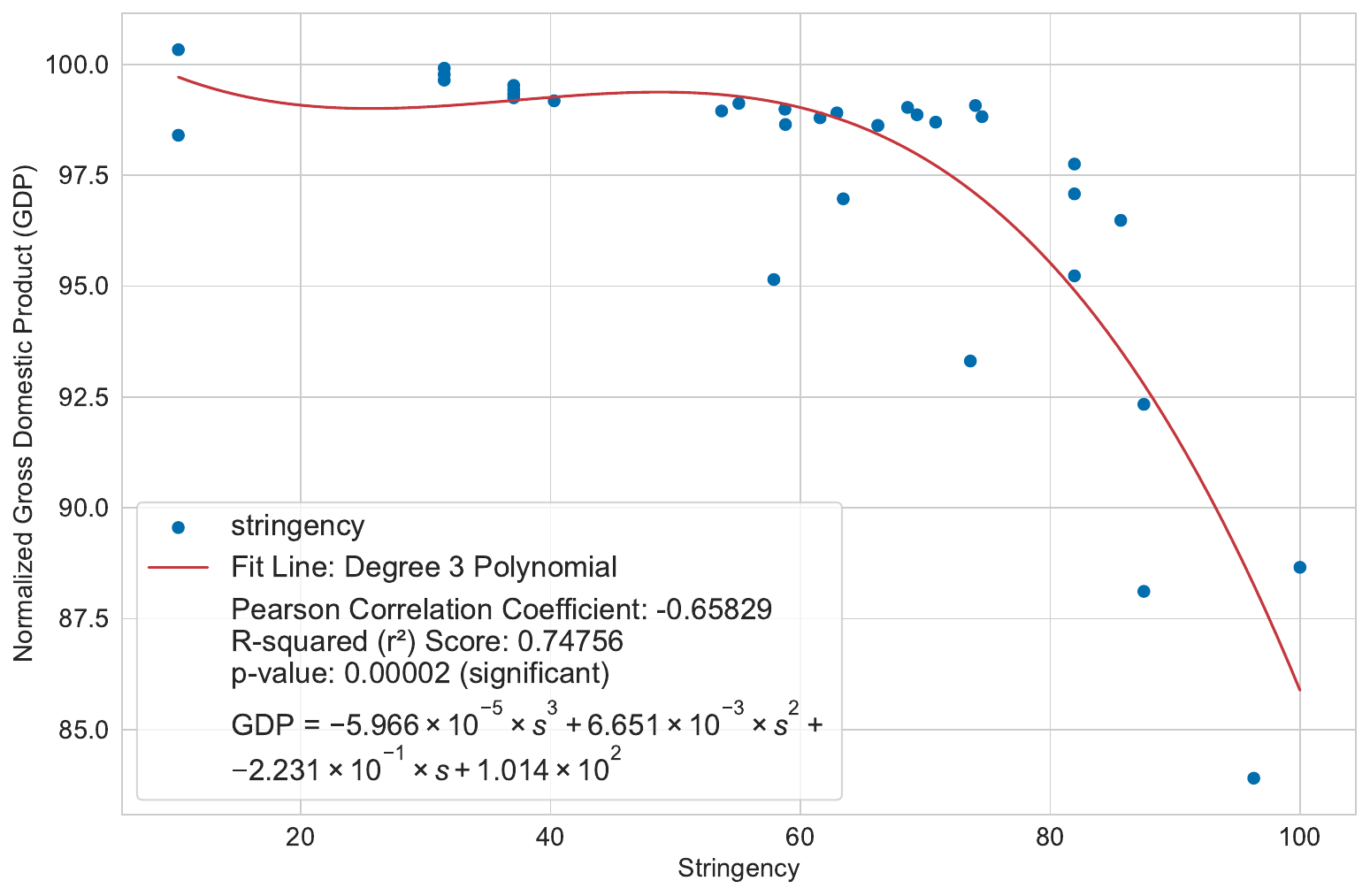}
    \caption{Stringency and Normalized GDP for India}
    \label{fig:stringency_vs_gdp_IND}
  \end{subfigure}
  \hfill
  \begin{subfigure}[t]{0.48\textwidth}
    \centering
    \includegraphics[width=\linewidth]{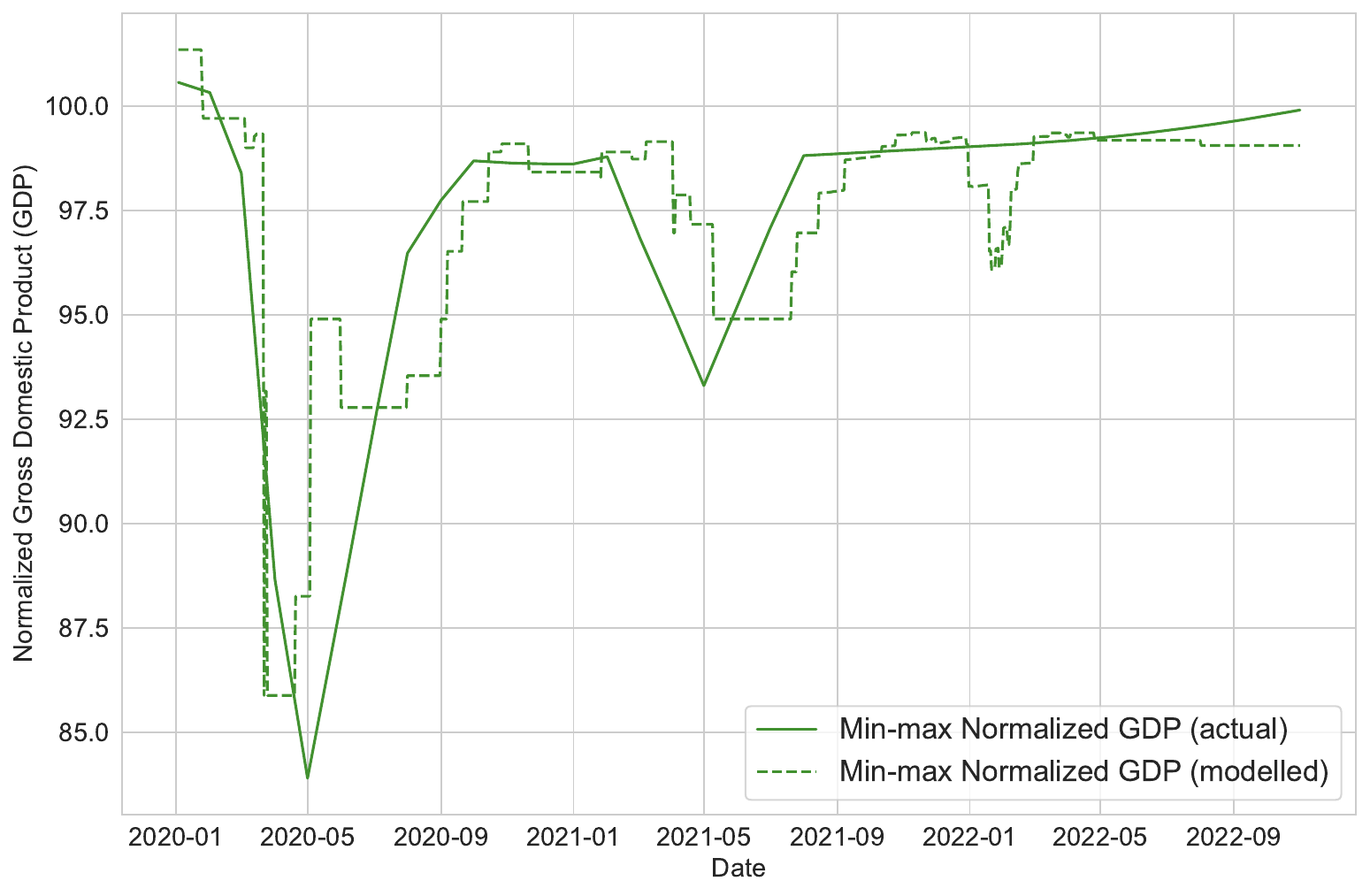}
    \caption{Normalized GDP modelled with Stringency for India}
    \label{fig:gdp_modelled_with_stringency_IND}
  \end{subfigure}
  \begin{subfigure}[t]{0.48\textwidth}
    \centering
    \includegraphics[width=\linewidth]{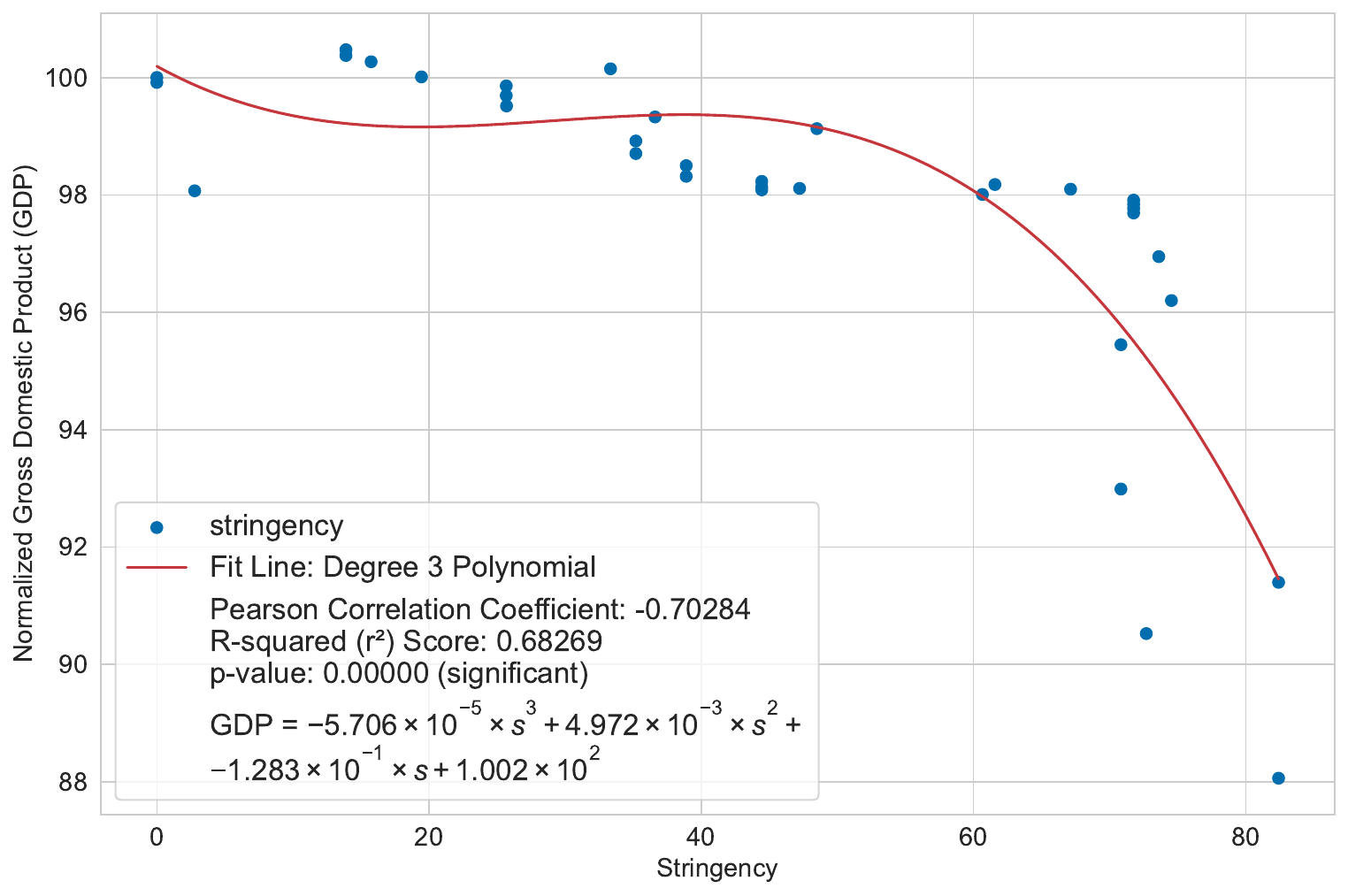}
    \caption{Stringency and Normalized GDP for Mexico}
    \label{fig:stringency_vs_gdp_MEX}
  \end{subfigure}
  \hfill
  \begin{subfigure}[t]{0.48\textwidth}
    \centering
    \includegraphics[width=\linewidth]{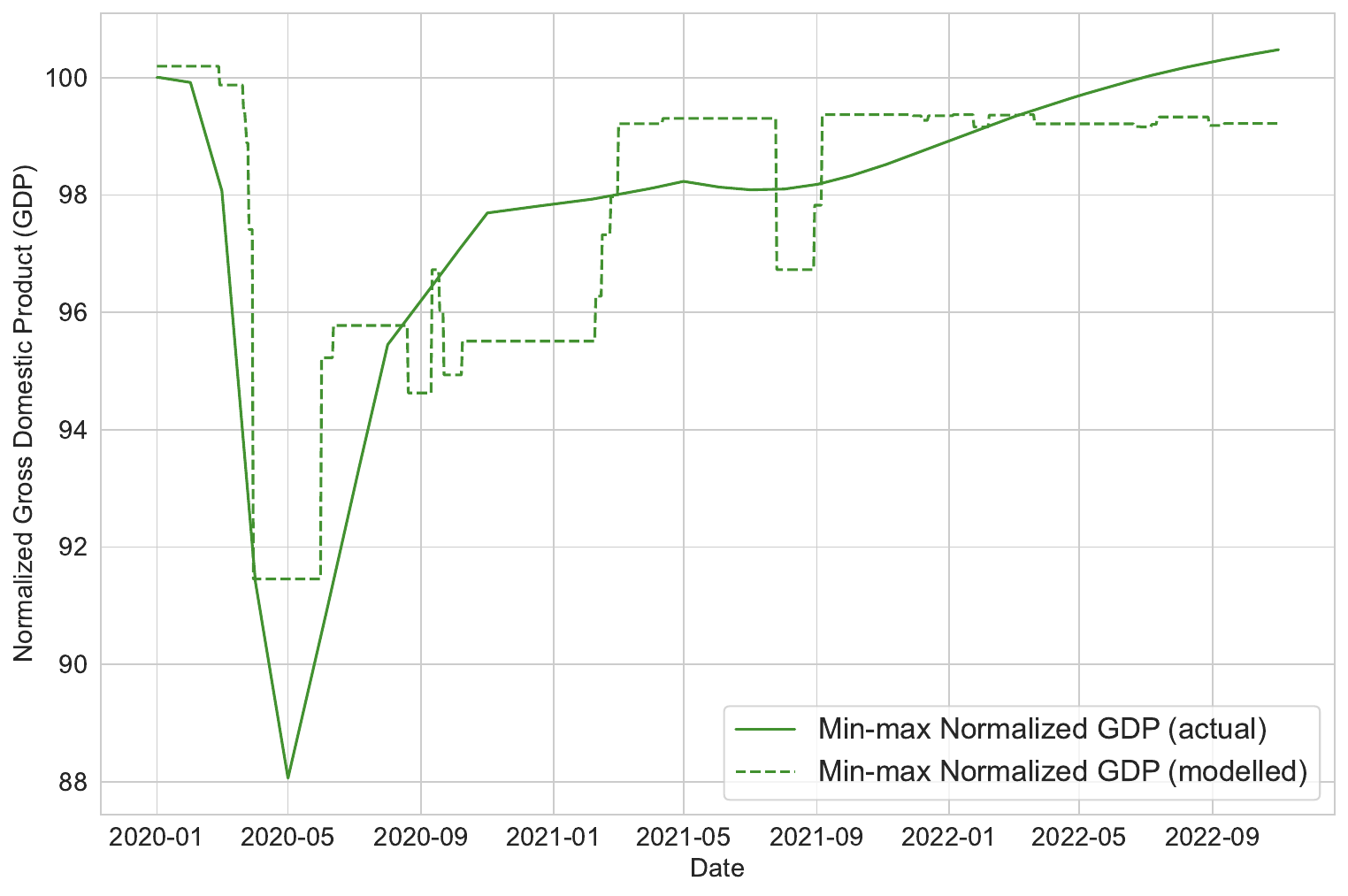}
    \caption{Normalized GDP modelled with Stringency for Mexico}
    \label{fig:gdp_modelled_with_stringency_MEX}
  \end{subfigure}
  \begin{subfigure}[t]{0.48\textwidth}
    \centering
    \includegraphics[width=\linewidth]{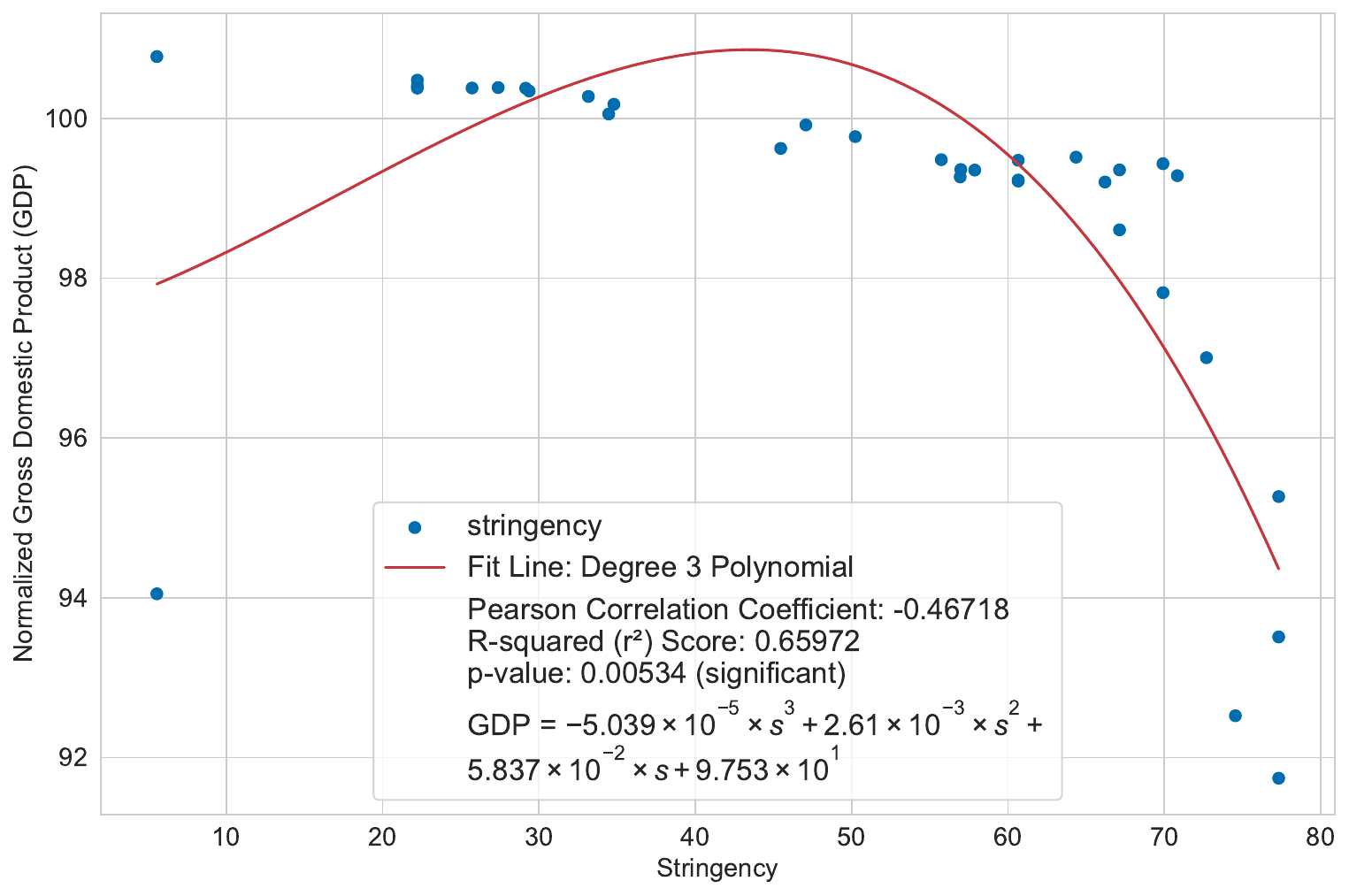}
    \caption{Stringency and Normalized GDP for Brazil}
    \label{fig:stringency_vs_gdp_BRA}
  \end{subfigure}
  \hfill
  \begin{subfigure}[t]{0.48\textwidth}
    \centering
    \includegraphics[width=\linewidth]{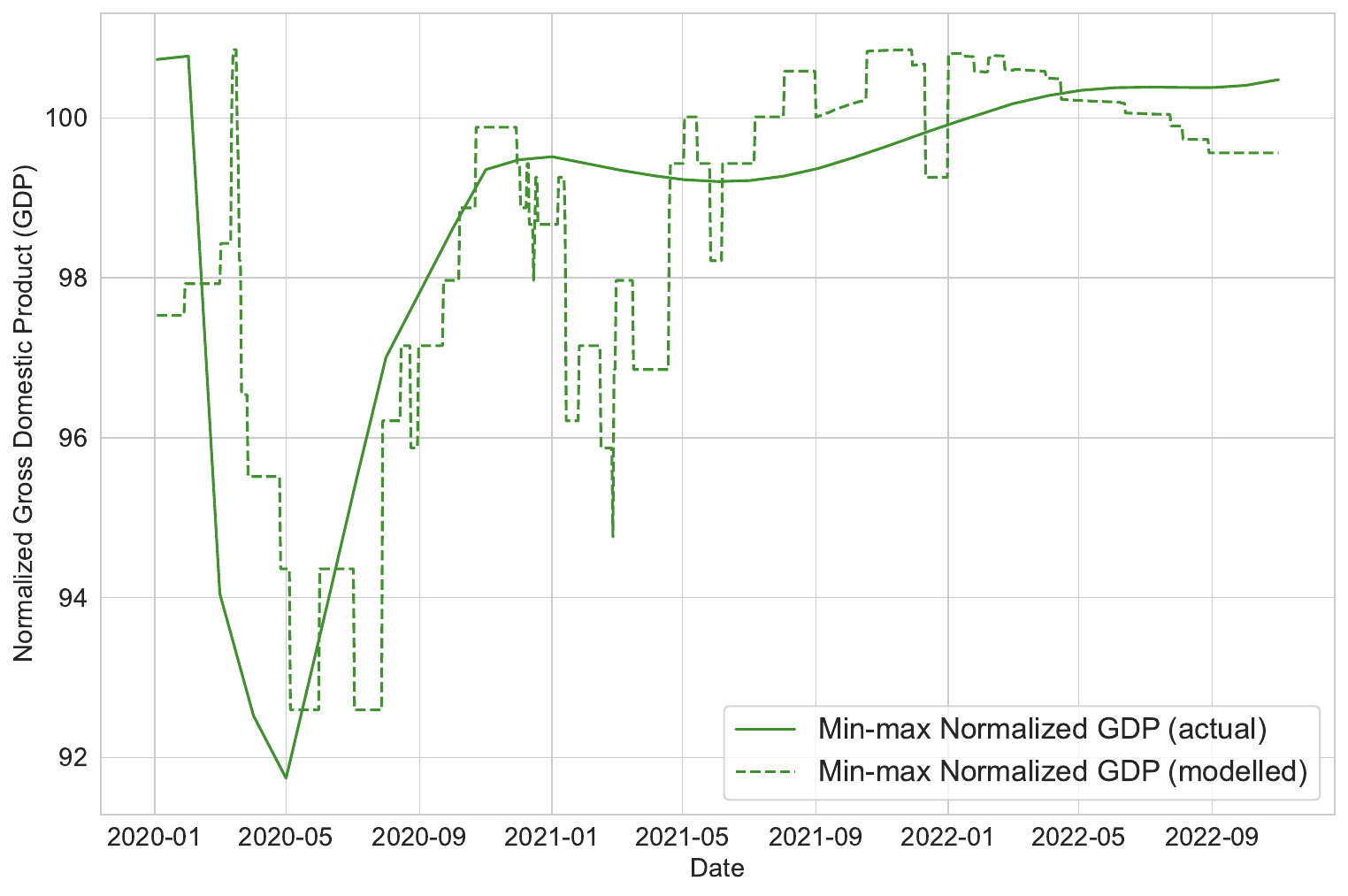}
    \caption{Normalized GDP modelled with Stringency for Brazil}
    \label{fig:gdp_modelled_with_stringency_BRA}
  \end{subfigure}
  \caption{\textbf{Stringency and GDP for Developing Economies.} Here, "(actual)" is the real data, "(modelled)" is the model for normalized GDP given the stringency. For countries with developing economies, when we model the normalized GDP with stringency we see significant p-values and high r2 scores.}
  \label{fig:stringency_gdp_developing}
\end{figure}

\begin{figure}[htbp!]
  \centering
  \begin{subfigure}[t]{0.48\textwidth}
    \centering
    \includegraphics[width=\linewidth]{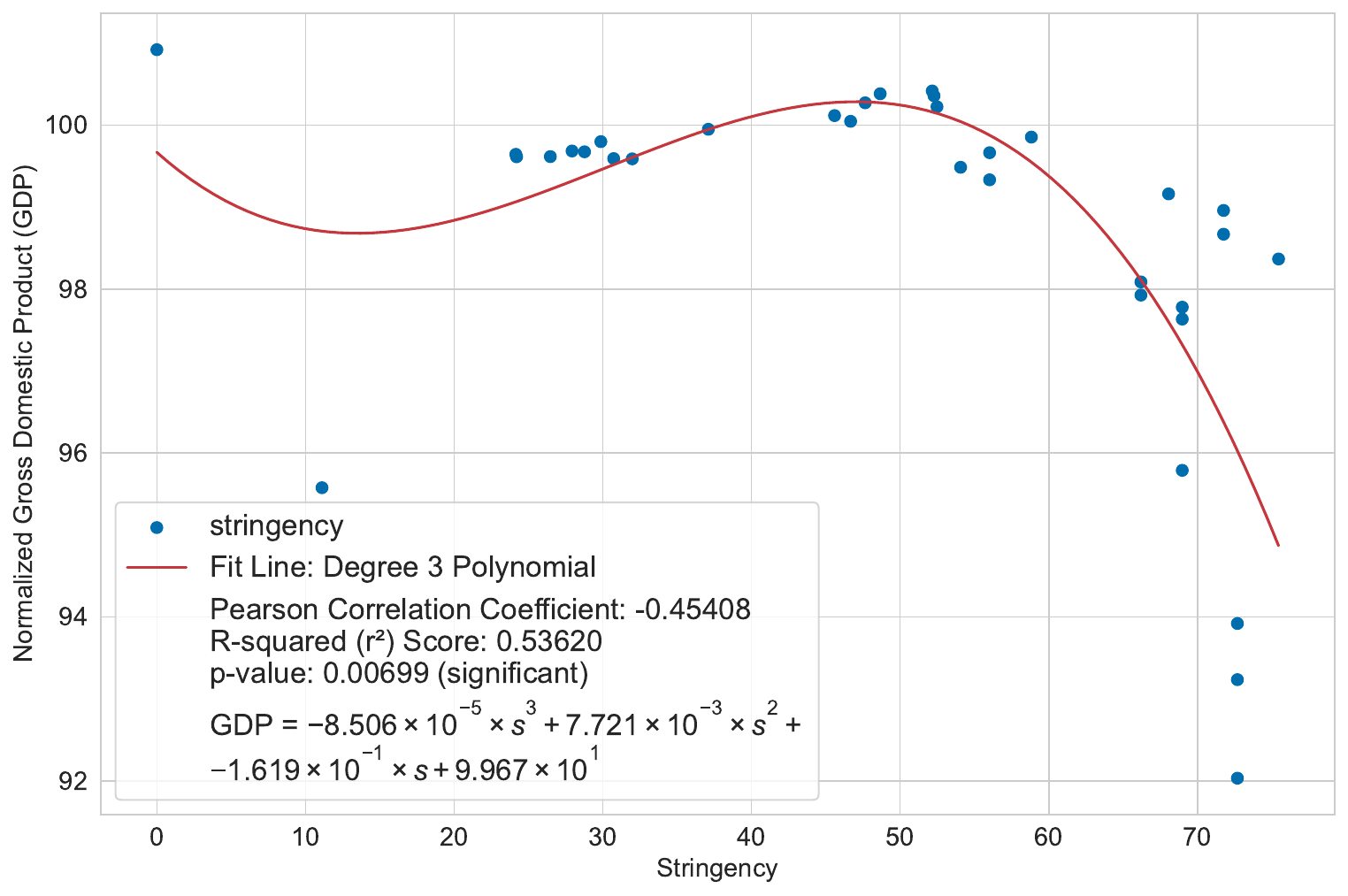}
    \caption{Stringency and Normalized GDP for United States}
    \label{fig:stringency_vs_gdp_USA}
  \end{subfigure}
  \hfill
  \begin{subfigure}[t]{0.48\textwidth}
    \centering
    \includegraphics[width=\linewidth]{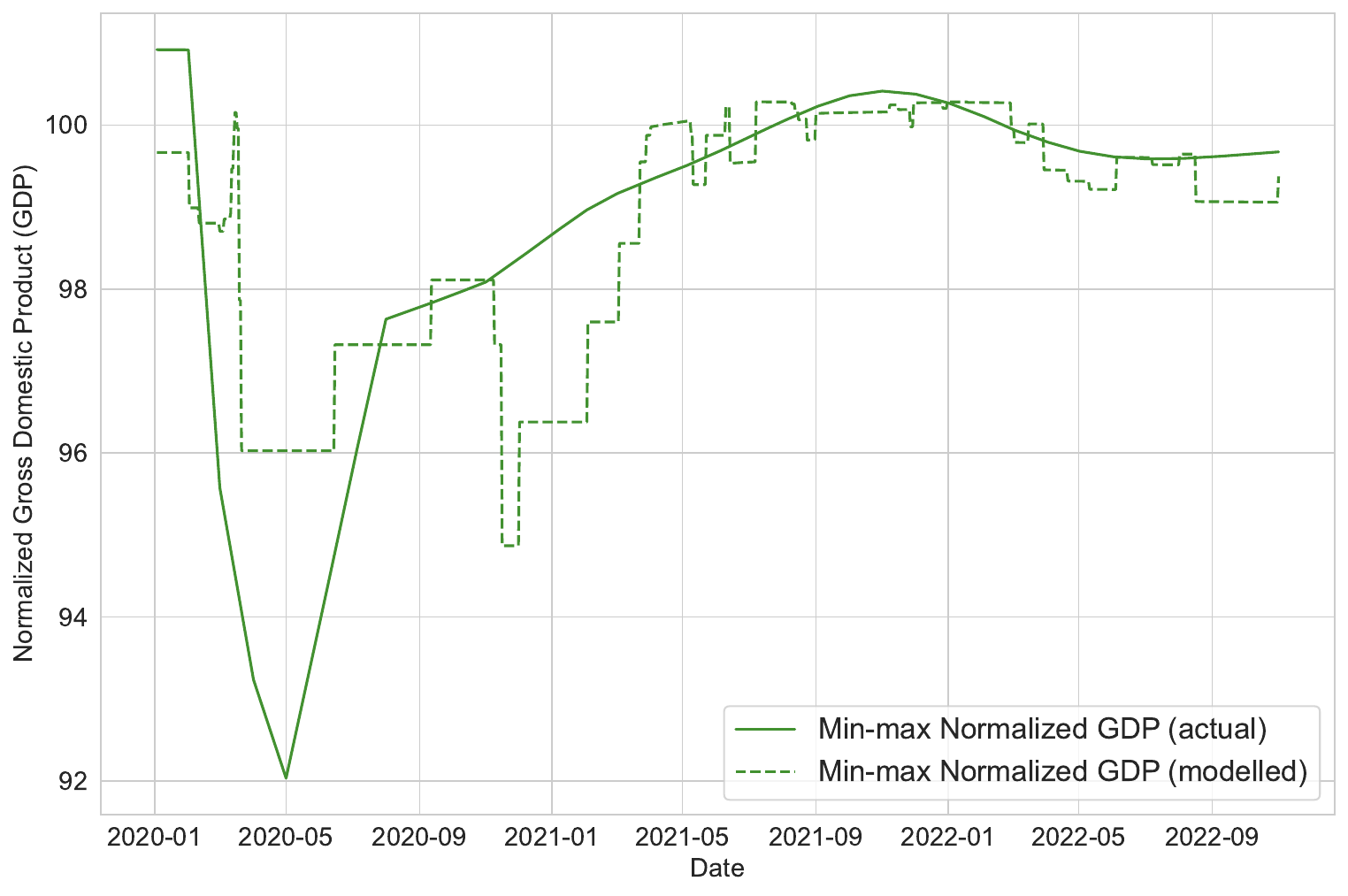}
    \caption{Normalized GDP modelled with Stringency for United States}
    \label{fig:gdp_modelled_with_stringency_USA}
  \end{subfigure}
  \begin{subfigure}[t]{0.48\textwidth}
    \centering
    \includegraphics[width=\linewidth]{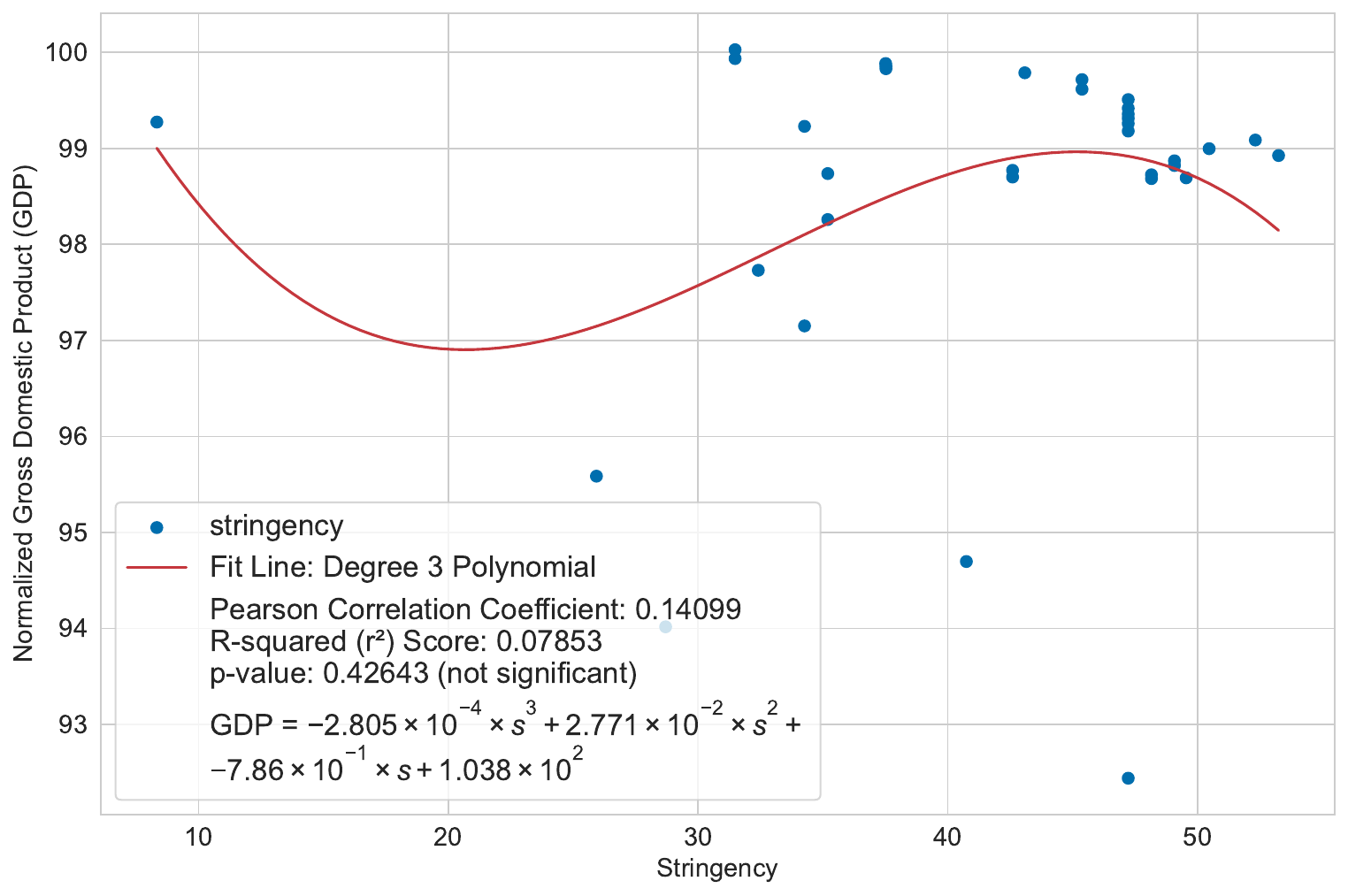}
    \caption{Stringency and Normalized GDP for Japan}
    \label{fig:stringency_vs_gdp_JPN}
  \end{subfigure}
  \hfill
  \begin{subfigure}[t]{0.48\textwidth}
    \centering
    \includegraphics[width=\linewidth]{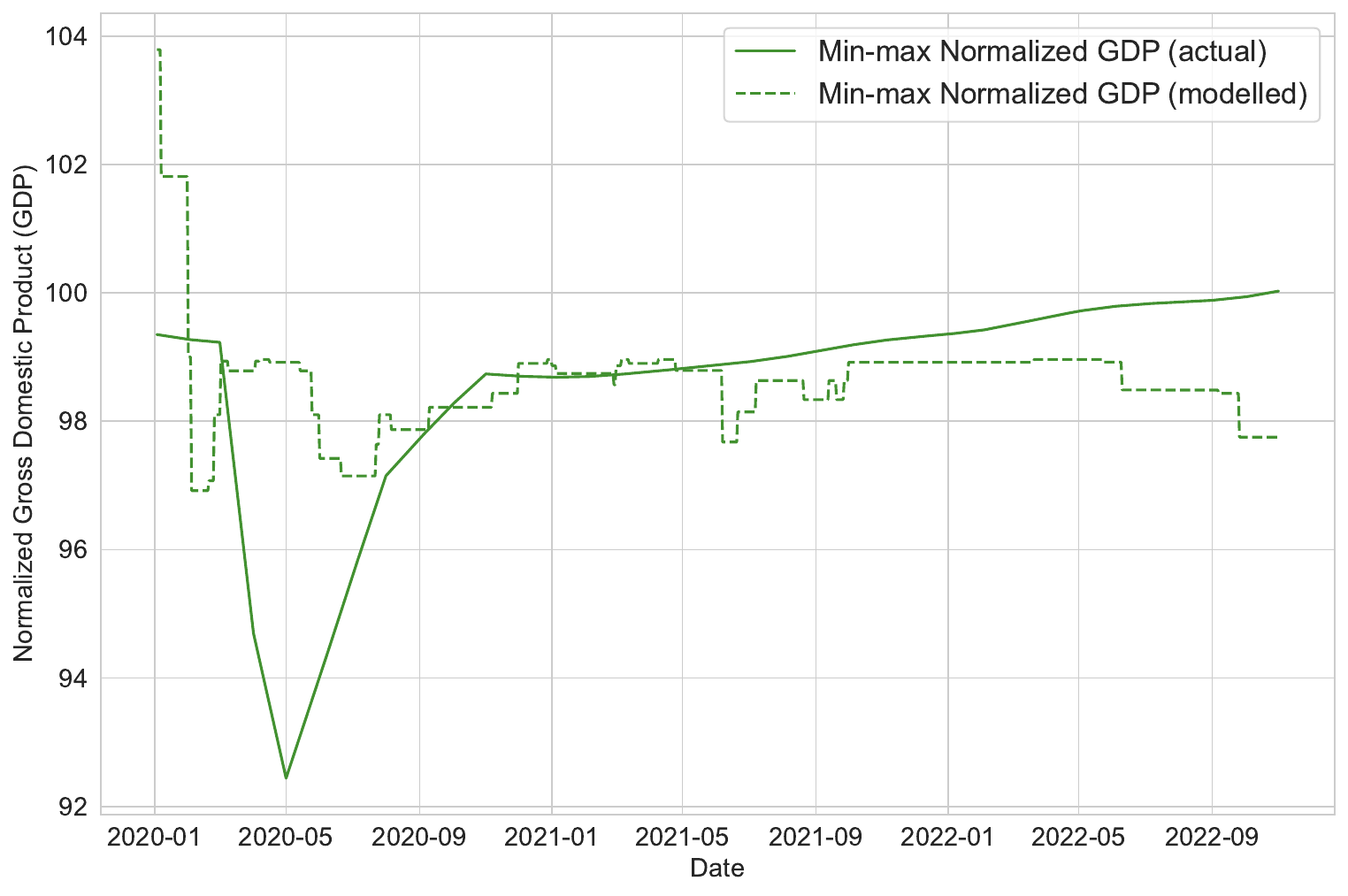}
    \caption{Normalized GDP modelled with Stringency for Japan}
    \label{fig:gdp_modelled_with_stringency_JPN}
  \end{subfigure}
  \begin{subfigure}[t]{0.48\textwidth}
    \centering
    \includegraphics[width=\linewidth]{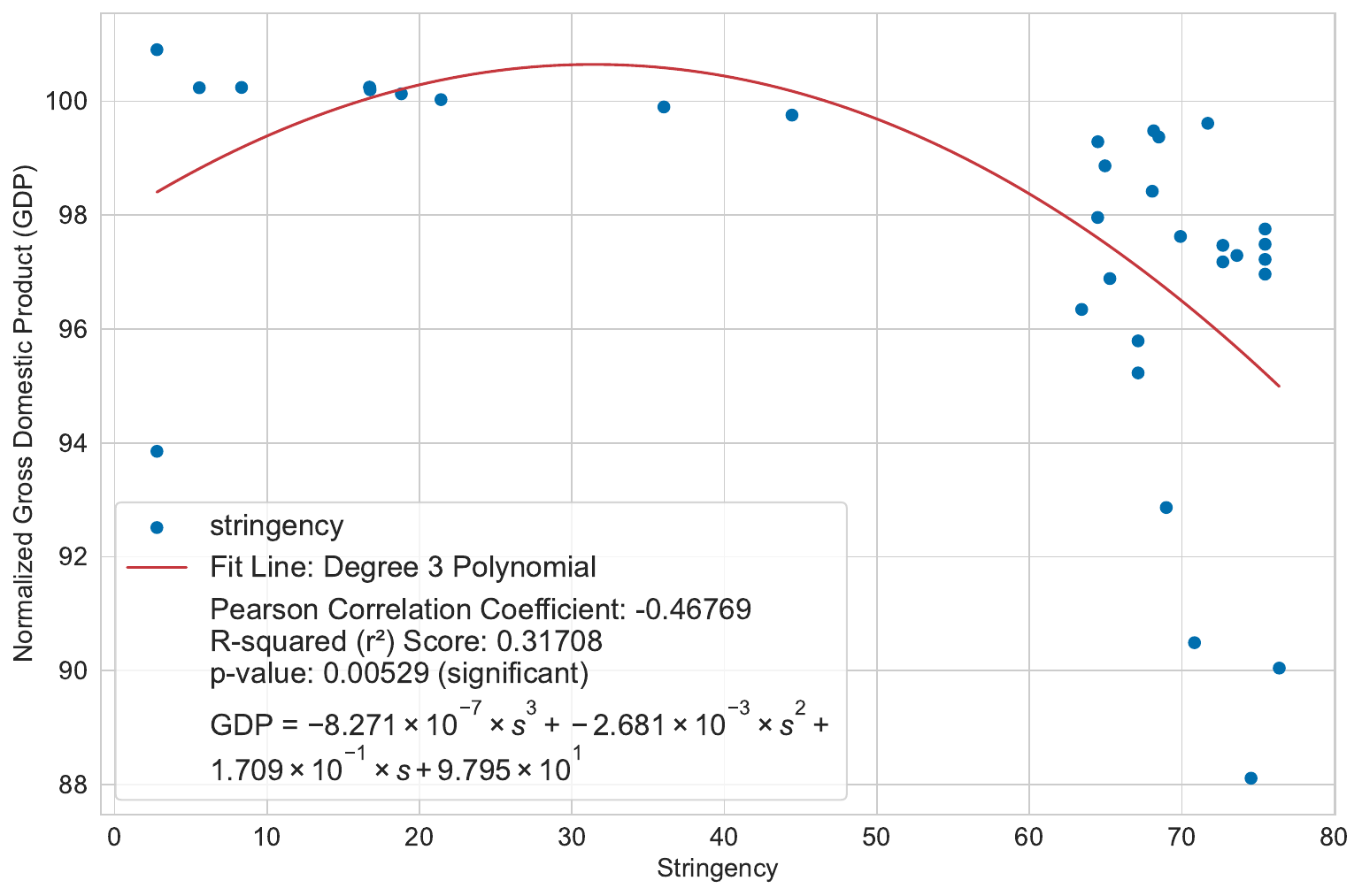}
    \caption{Stringency and Normalized GDP for Canada}
    \label{fig:stringency_vs_gdp_CAN}
  \end{subfigure}
  \hfill
  \begin{subfigure}[t]{0.48\textwidth}
    \centering
    \includegraphics[width=\linewidth]{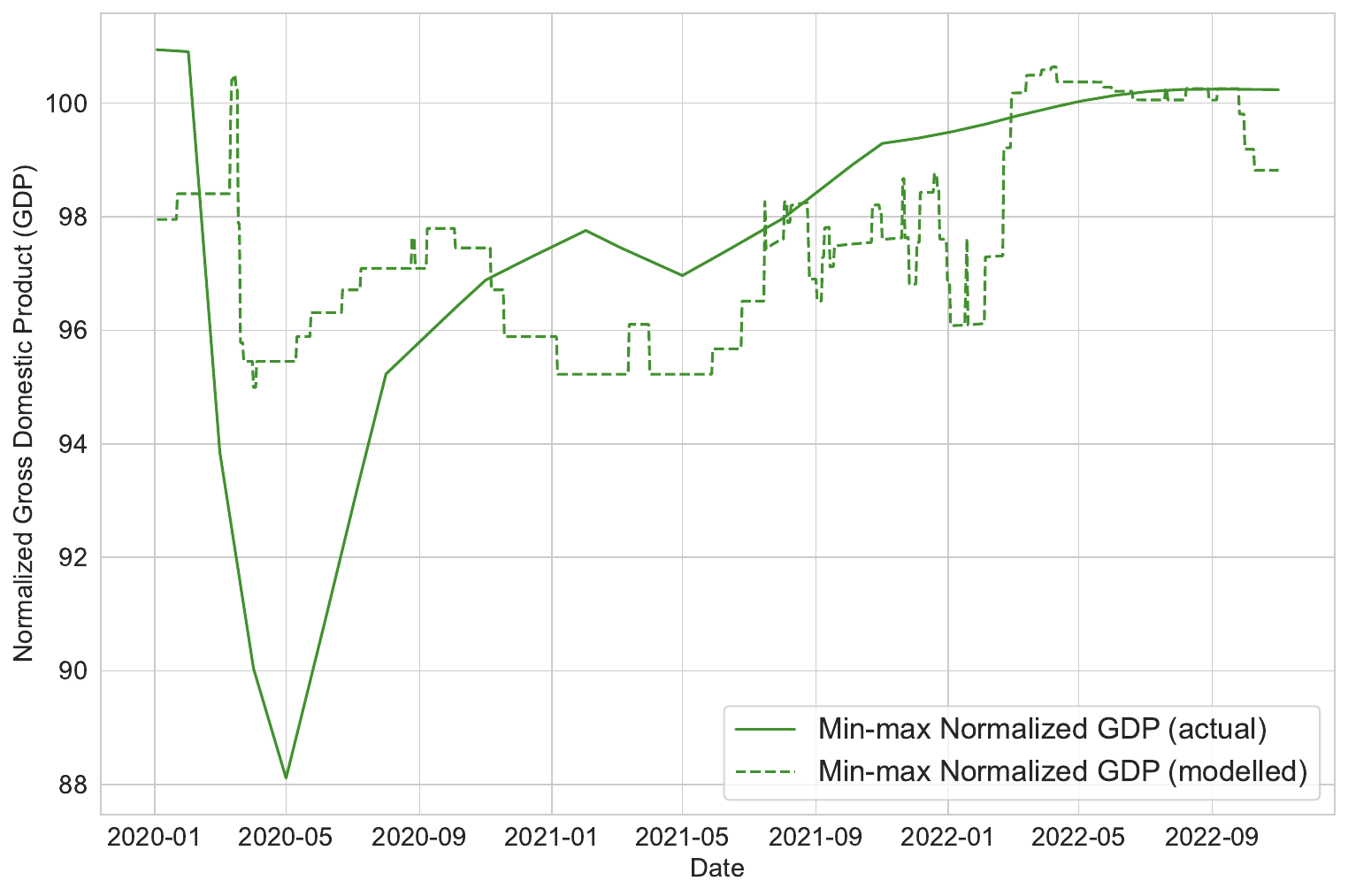}
    \caption{Normalized GDP modelled with Stringency for Canada}
    \label{fig:gdp_modelled_with_stringency_CAN}
  \end{subfigure}
  \caption{\textbf{Stringency and GDP for Advanced Economies.} Here, "(actual)" is the real data, "(modelled)" is the model for normalized GDP given the stringency. In economically advanced nations, when modeling the normalized GDP against stringency measures, we observe substantial p-values, providing evidence against the null hypothesis. However, the significance levels are not as high as those found for developing economies (\crefrange{fig:stringency_vs_gdp_IND}{fig:gdp_modelled_with_stringency_BRA}). This is reflected in the lower R-squared scores, which indicates that the relationship between these variables may be less pronounced when compared to developing economies.}
  \label{fig:stringency_gdp_advanced}
\end{figure}

While non-pharmaceutical interventions (NPIs) can effectively manage the epidemic, they impose economic burdens on developing nations. In \cref{fig:stringency_gdp_developing}, we plot the normalized GDP against the stringency and calculate various metrics like the Pearson correlation coefficient, coefficient of determination ($r^2$), and p-value for three countries (India, Mexico, Brazil), which are Emerging Market and Developing Economies~\cite{IMFCovid} from May 2020 to October 2022. It can be observed from \cref{fig:stringency_gdp_developing} that strict policies have a negative effect on the normalized GDP in these economies. However, this trend is not uniformly seen in advanced economies like the USA, Japan and Canada as shown in \cref{fig:stringency_gdp_advanced}. In these countries, other factors could be contributing to the decrease in normalized GDP besides the implementation of stricter policies.

\begin{figure}[htbp!]
  \centering
  \begin{subfigure}[t]{0.48\textwidth}
    \centering
    \includegraphics[width=\linewidth]{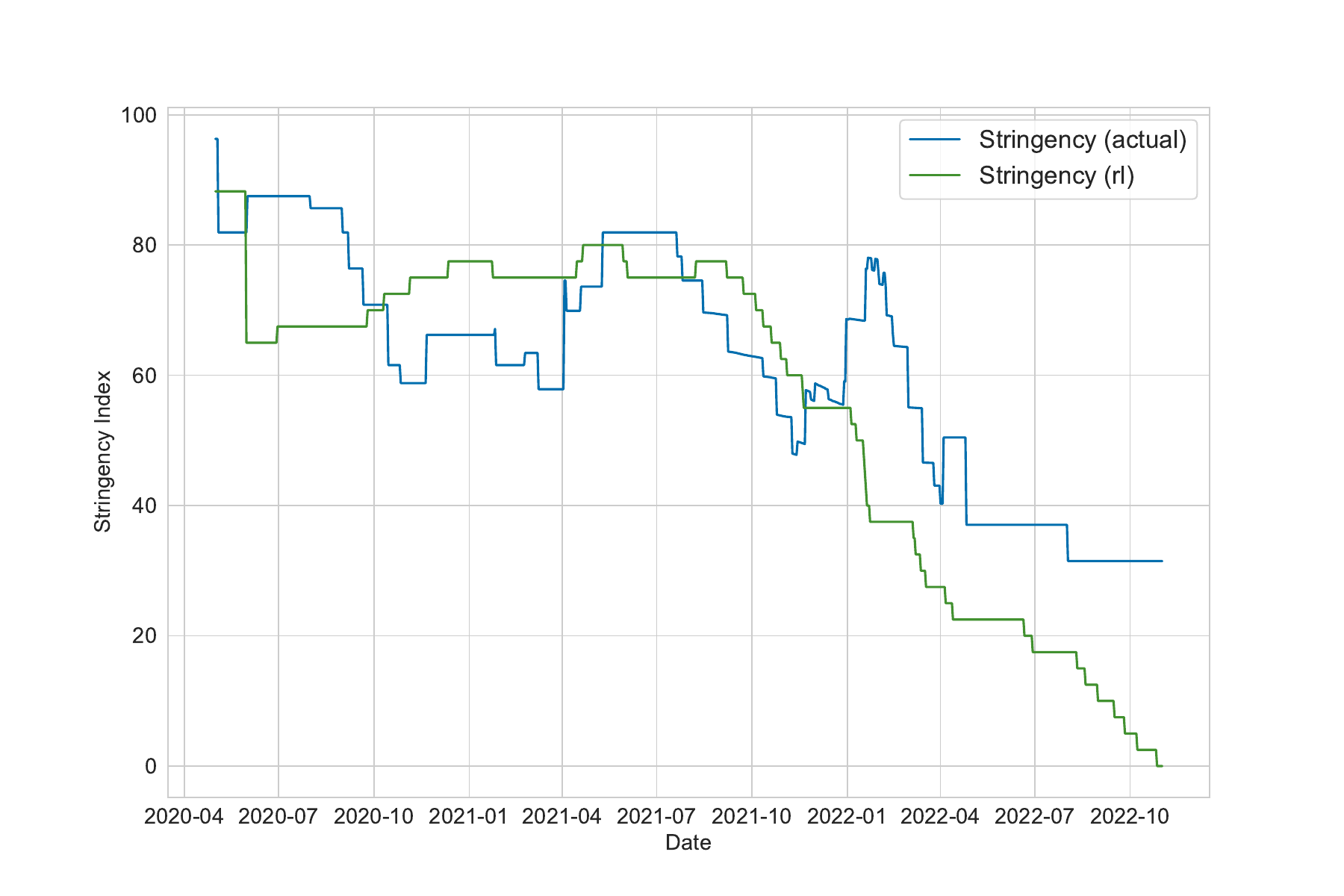}
    \caption{Stringency changing over Time}
    \label{fig:176647_rl_stringency}
  \end{subfigure}
  \hfill
  \begin{subfigure}[t]{0.48\textwidth}
    \centering
    \includegraphics[width=\linewidth]{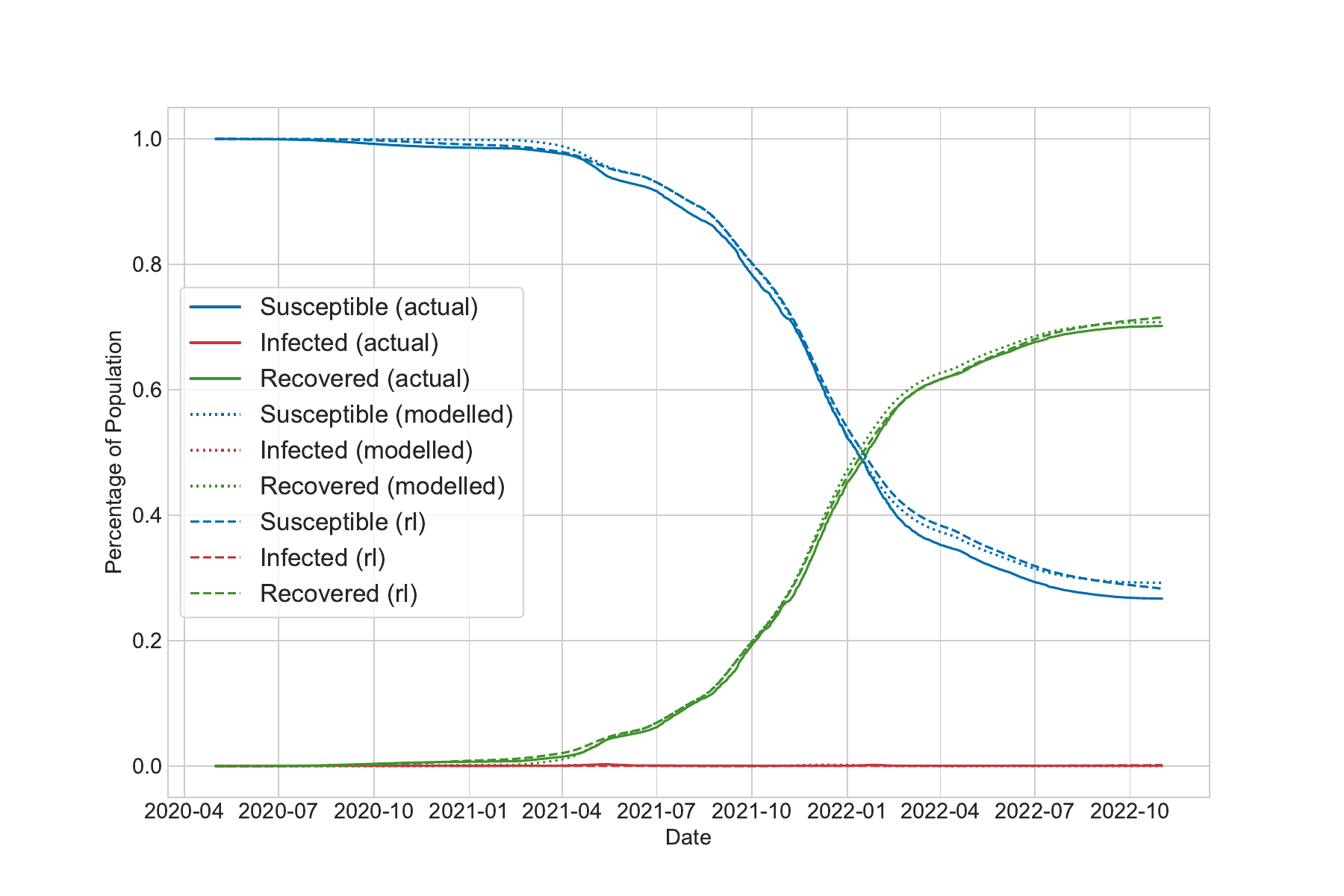}
    \caption{SIR Dynamics}
    \label{fig:176647_rl_sir}
  \end{subfigure}
  \hfill
  \begin{subfigure}[t]{0.48\textwidth}
    \centering
    \includegraphics[width=\linewidth]{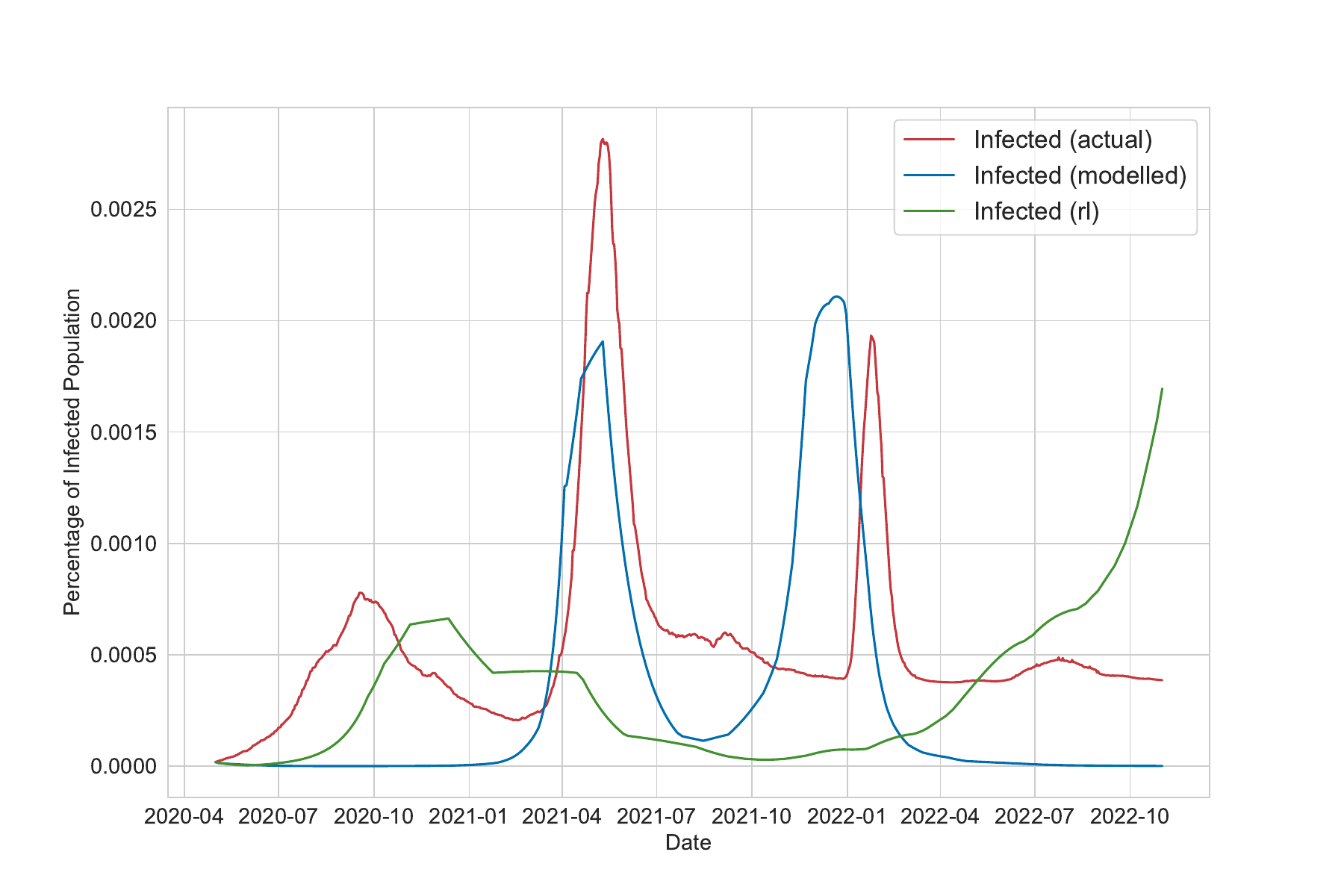}
    \caption{Infected Population changing over Time}
    \label{fig:176647_rl_i}
  \end{subfigure}
  \hfill
  \begin{subfigure}[t]{0.48\textwidth}
    \centering
    \includegraphics[width=\linewidth]{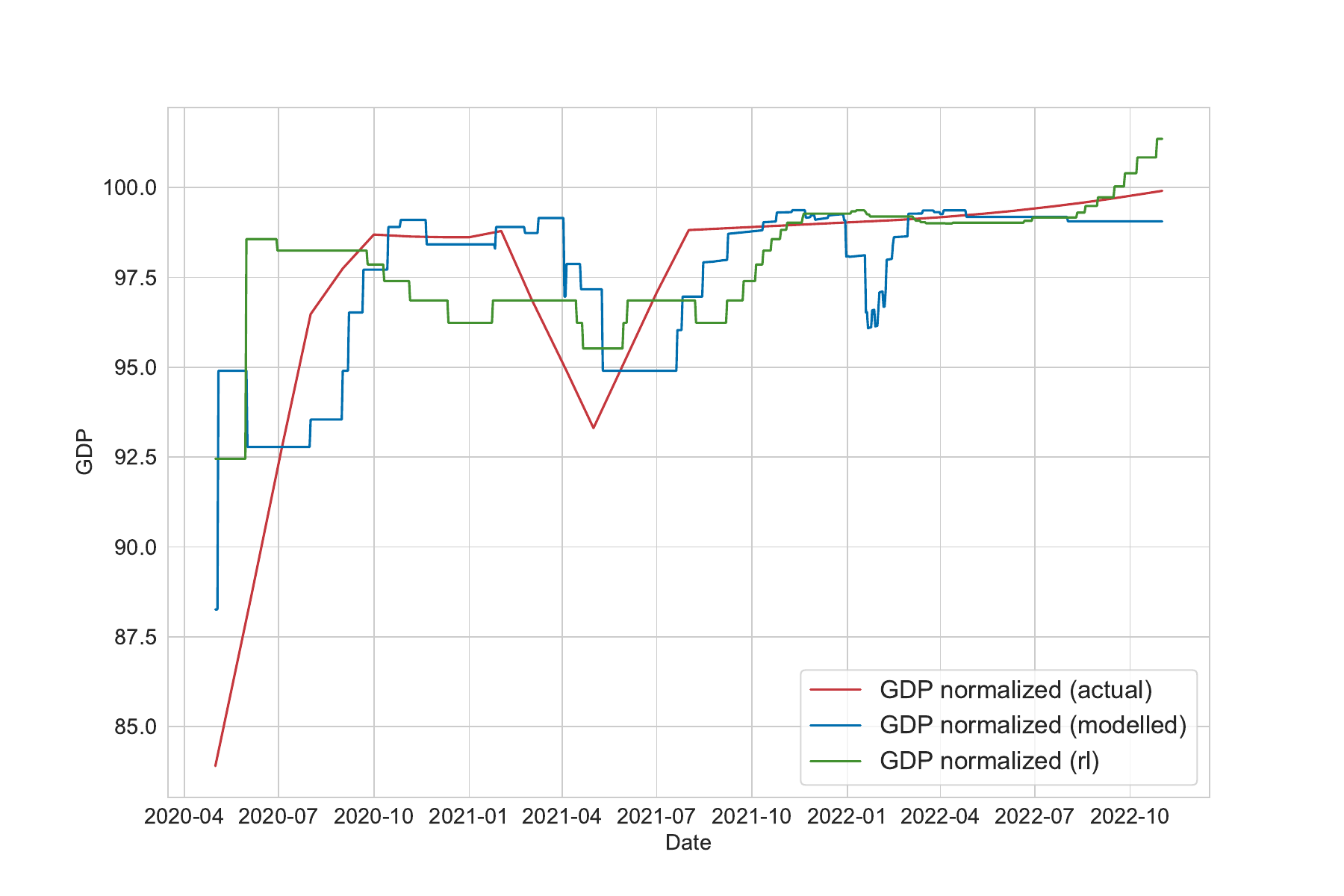}
    \caption{Normalized GDP changing over Time}
    \label{fig:176647_rl_gdp}
  \end{subfigure}
  \hfill
  \begin{subfigure}[t]{0.48\textwidth}
    \centering
    \includegraphics[width=\linewidth]{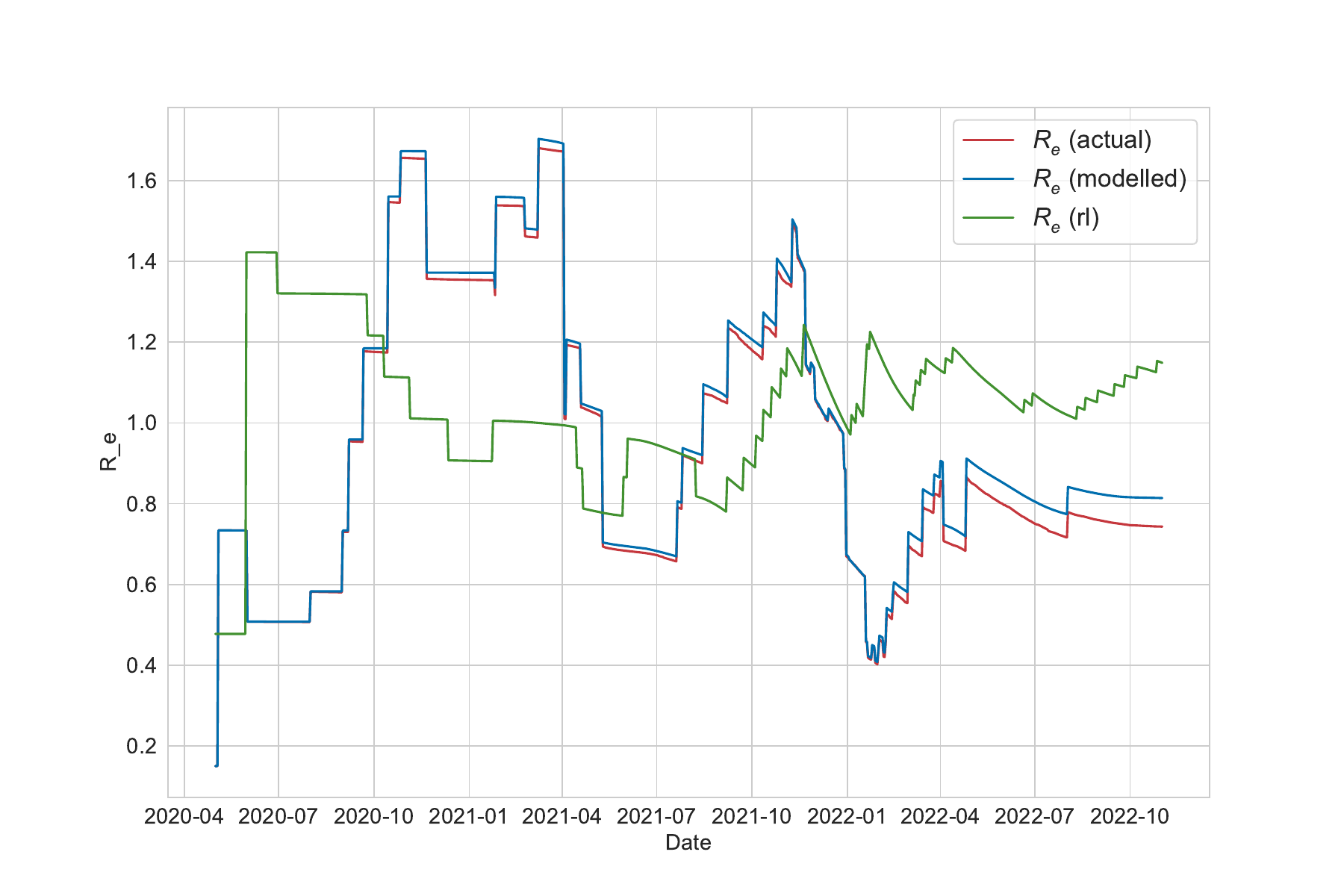}
    \caption{$R_e$ changing over Time}
    \label{fig:176647_rl_r_eff}
  \end{subfigure}
  \hfill
  \begin{subfigure}[t]{0.48\textwidth}
    \centering
    \includegraphics[width=\linewidth]{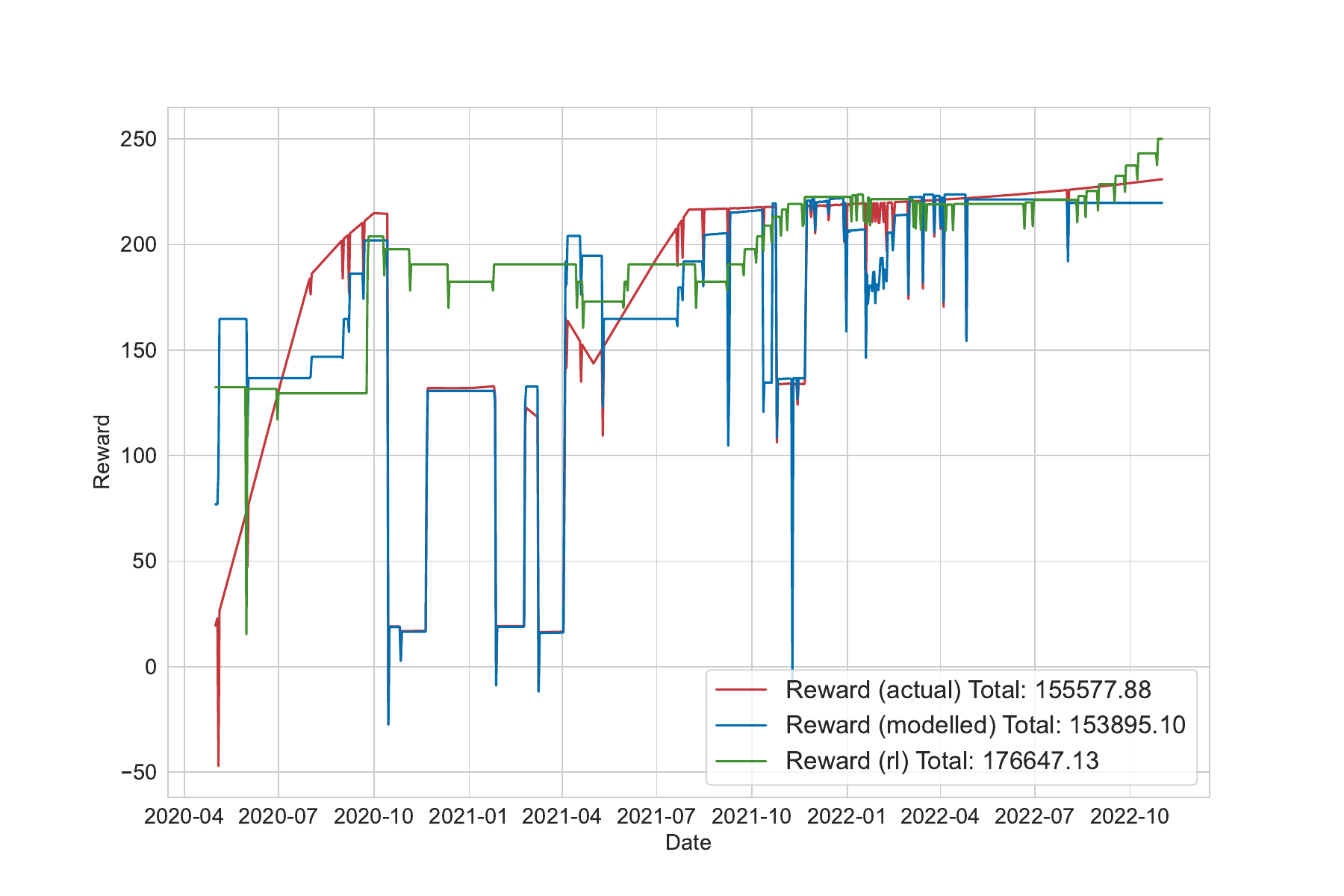}
    \caption{Reward changing over Time}
    \label{fig:176647_rl_reward}
  \end{subfigure}
  \caption{\textbf{Strategy from Reinforcement Learning Agent.} Here, "(actual)" is the real data, "(modelled)" is the result from real world stringency imposed with the use of the SIR model with lockdown and time-varying vaccination rate, and "(rl)" is the new stringency strategy we propose. \textbf{(a)} The new strategy proposed highlights a decrease in stringency from July, 2020 till October, 2020 compared to the actual data. After October, 2020 there's an increase and then a steady decline towards the end. \textbf{(c)} There's a peak in the number of infected people around October, 2020 and then a second peak after October, 2022 \textbf{(d)} The normalized GDP is also maintained and doesn't show a dip during April, 2022. \textbf{(e)} The $R_e$ is maintained below $1.5$ throughout, and below $1.2$ after October, 2020. \textbf{(f)} A higher reward is achieved by the reinforcement learning agent.}
  \label{fig:176647_parent}
\end{figure}

After median filtering to smooth the output (to reinforce the negative reward from changing the stringencies) from the trained reinforcement learning agent, here are some of the results obtained. In the presented result \cref{fig:176647_parent}, we can see the reinforcement learning agent outperform the modelled outcome. A strategic decision is made by the agent to maintain the stringency index below 80 after April, 2020 \cref{fig:176647_rl_stringency}. This approach allows for the natural progression of disease dynamics, resulting in a rapid reduction of the effective reproduction number $R_e$ to below $1.2$ after October, 2020 (refer to figure \cref{fig:176647_rl_r_eff}). After October, 2021 there's a decrease in the stringency which leads to an increase in the normalized GDP, indicating an economic upturn. While this strategy poses a higher risk in terms of infection rates during the initial phase of the epidemic (prior to vaccine rollout) as well as the later phase (second peak of infected individuals \cref{fig:176647_rl_i}), but it proves to be more beneficial for the nation's economy in the long run. Despite the economic benefits in the long run, this strategy is not the most effective for the government to adopt due to the high number of infected individuals. Therefore, we propose an alternative strategy that involves some loss in terms of the economic impact.

\begin{figure}[htbp!]
  \centering
  \begin{subfigure}[t]{0.48\textwidth}
    \centering
    \includegraphics[width=\linewidth]{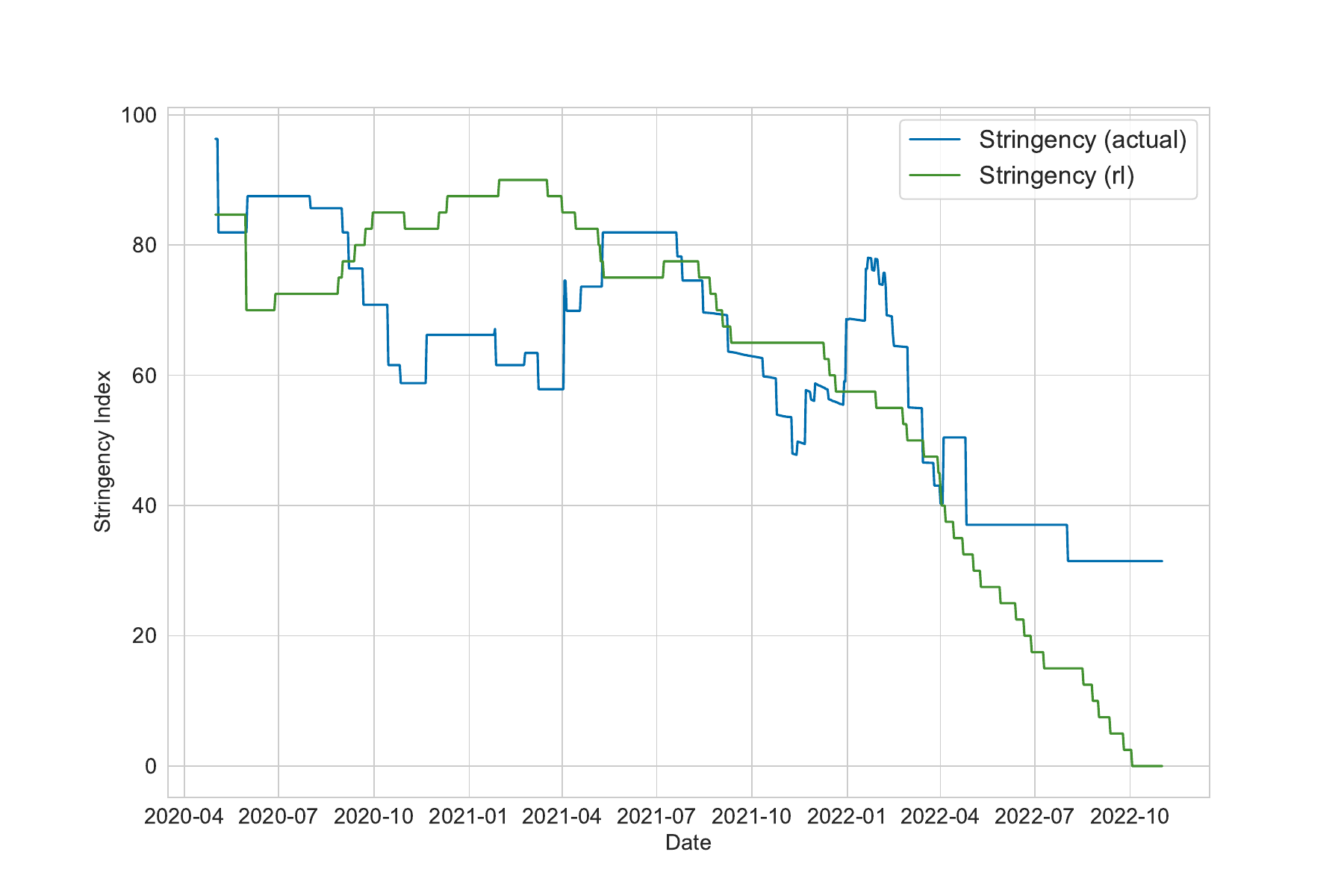}
    \caption{Stringency changing over Time}
    \label{fig:175975_rl_stringency}
  \end{subfigure}
  \hfill
  \begin{subfigure}[t]{0.48\textwidth}
    \centering
    \includegraphics[width=\linewidth]{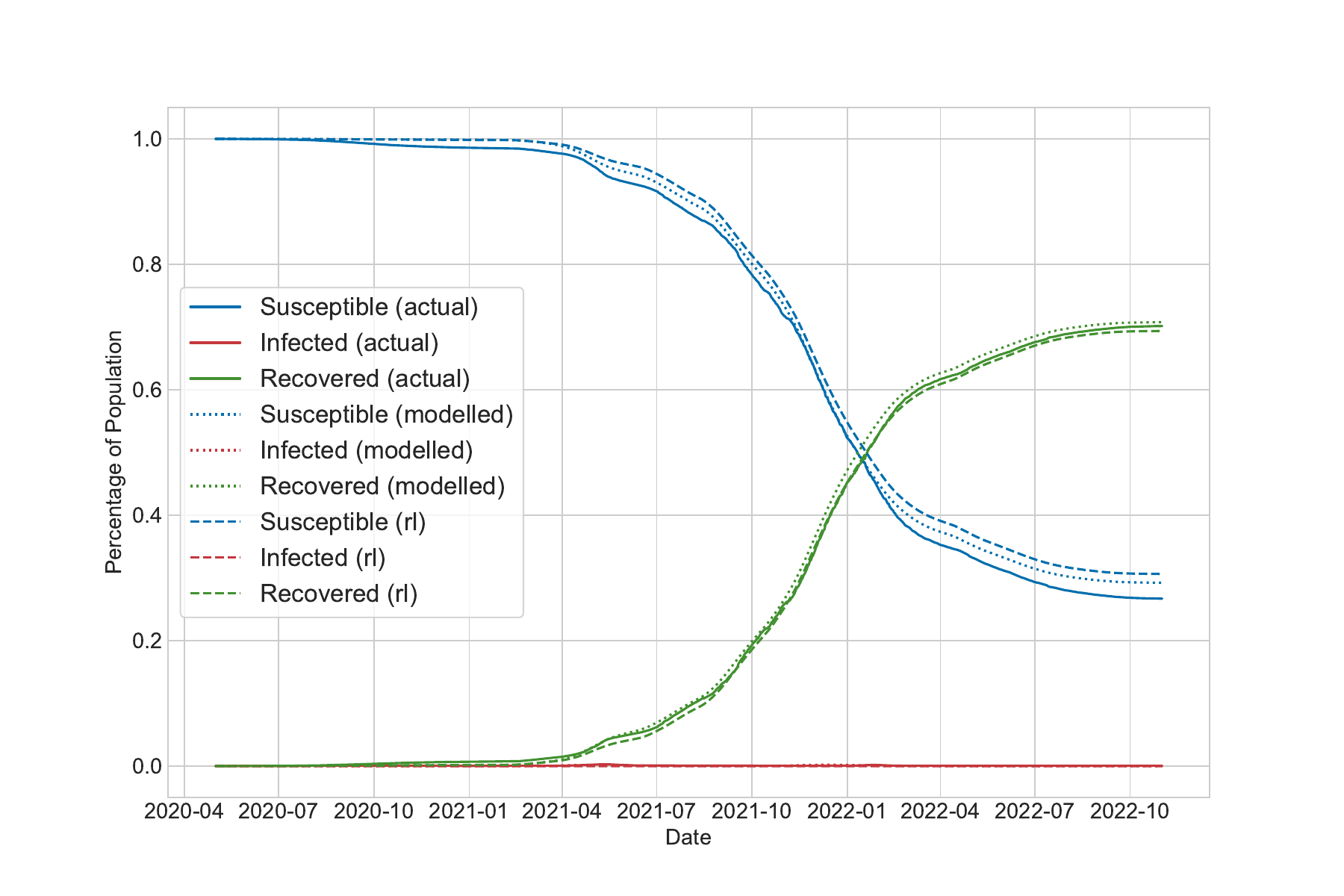}
    \caption{SIR Dynamics}
    \label{fig:175975_rl_sir}
  \end{subfigure}
  \hfill
  \begin{subfigure}[t]{0.48\textwidth}
    \centering
    \includegraphics[width=\linewidth]{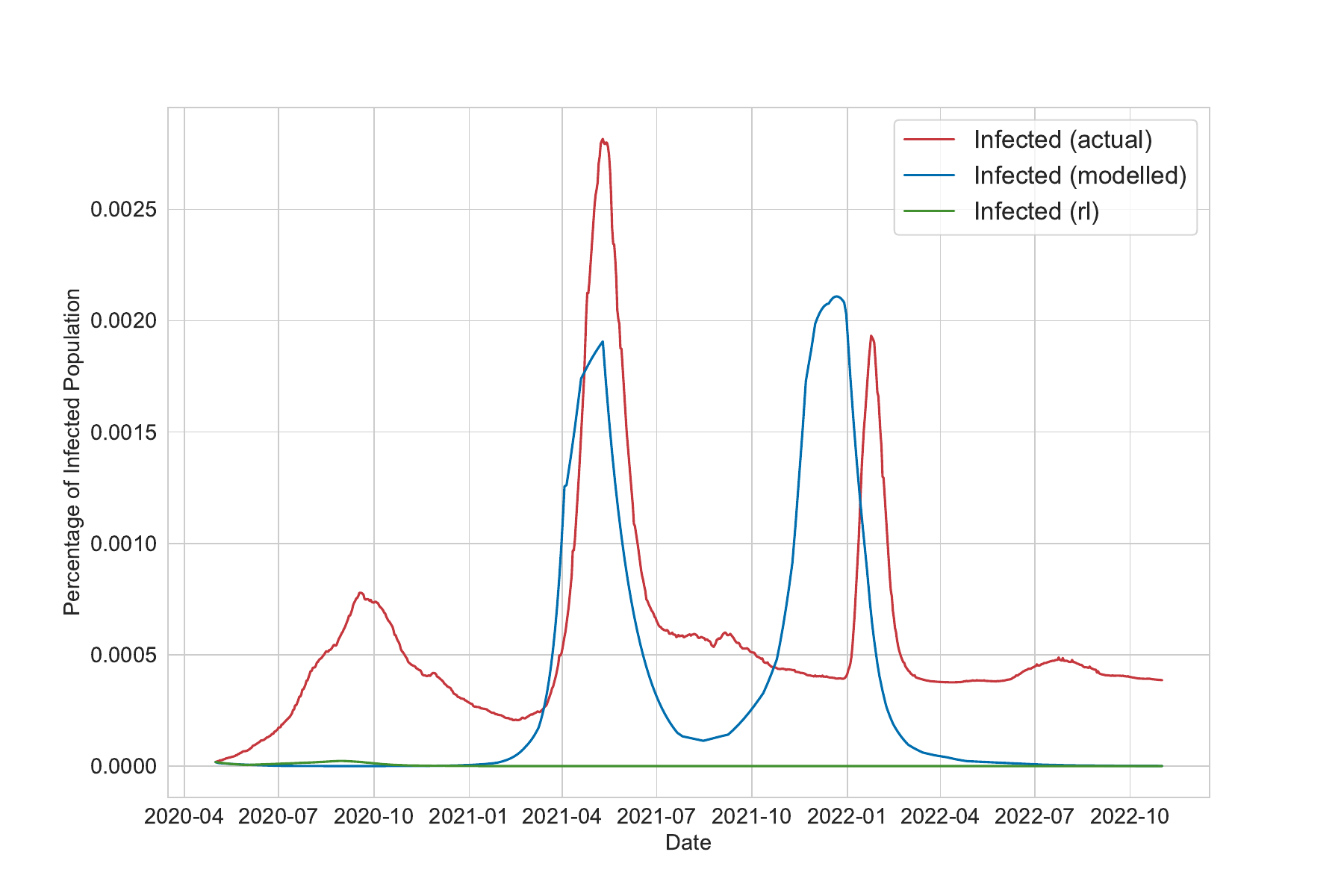}
    \caption{Infected Population changing over Time}
    \label{fig:175975_rl_i}
  \end{subfigure}
  \hfill
  \begin{subfigure}[t]{0.48\textwidth}
    \centering
    \includegraphics[width=\linewidth]{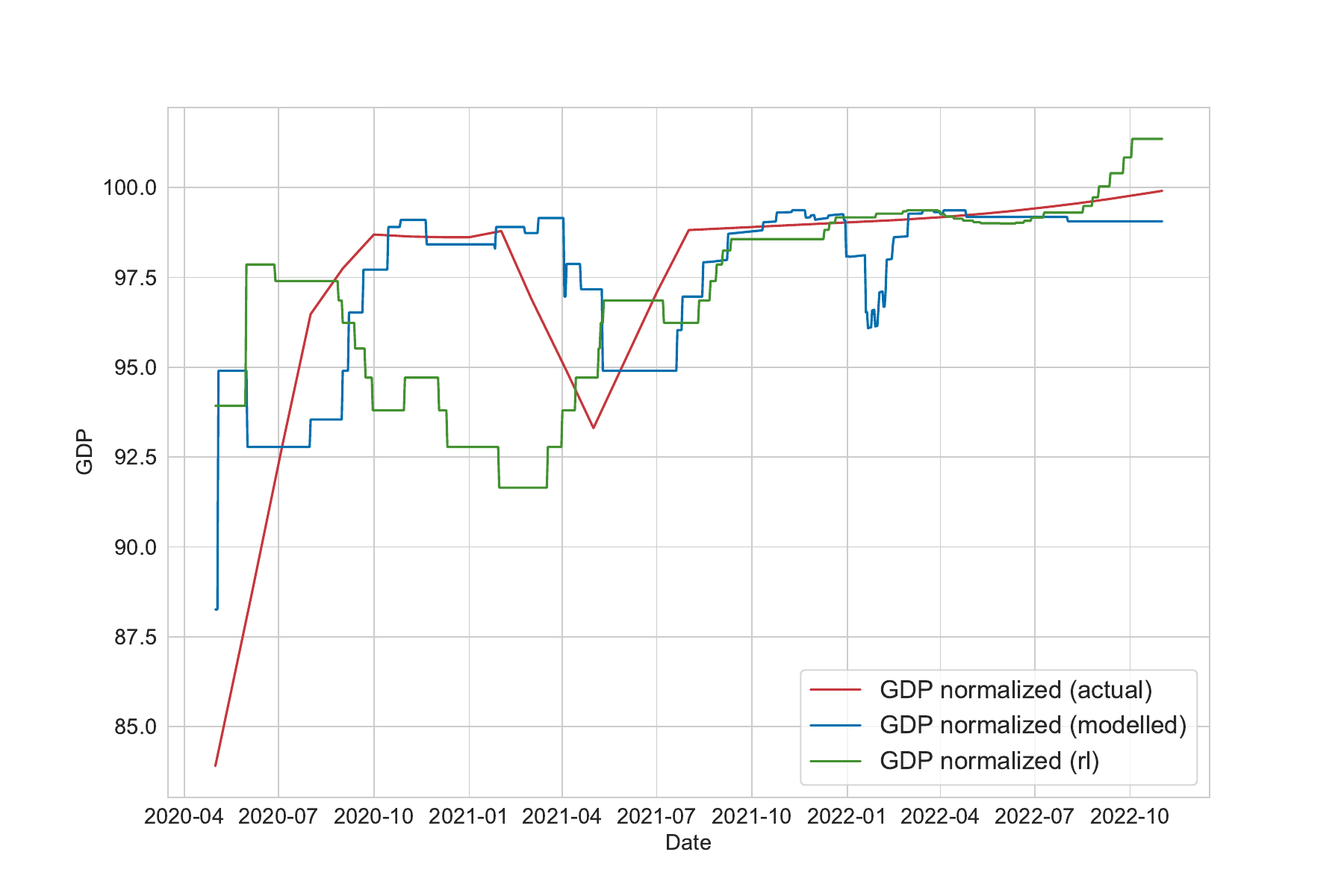}
    \caption{Normalized GDP changing over Time}
    \label{fig:175975_rl_gdp}
  \end{subfigure}
  \hfill
  \begin{subfigure}[t]{0.48\textwidth}
    \centering
    \includegraphics[width=\linewidth]{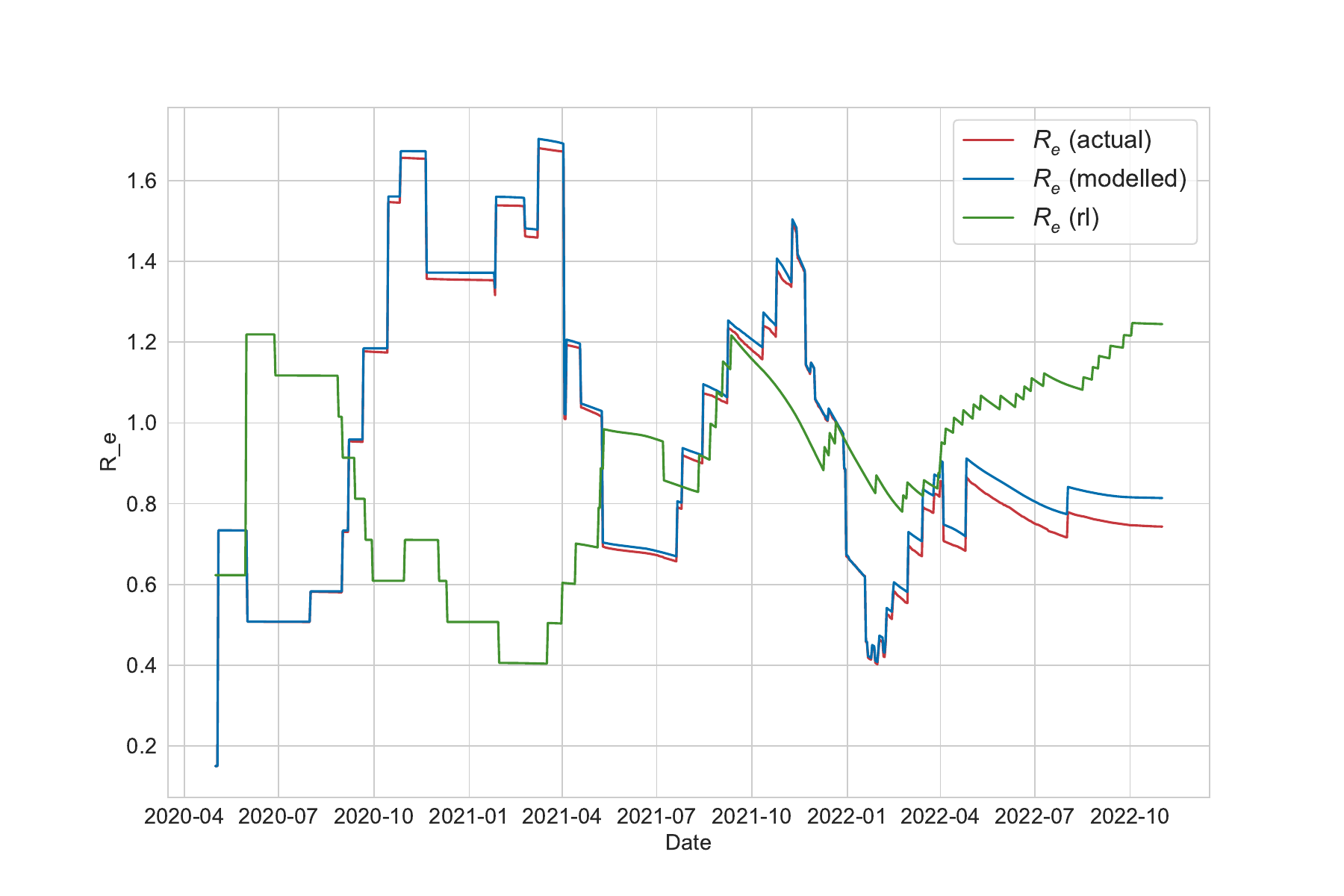}
    \caption{$R_e$ changing over Time}
    \label{fig:175975_rl_r_eff}
  \end{subfigure}
  \hfill
  \begin{subfigure}[t]{0.48\textwidth}
    \centering
    \includegraphics[width=\linewidth]{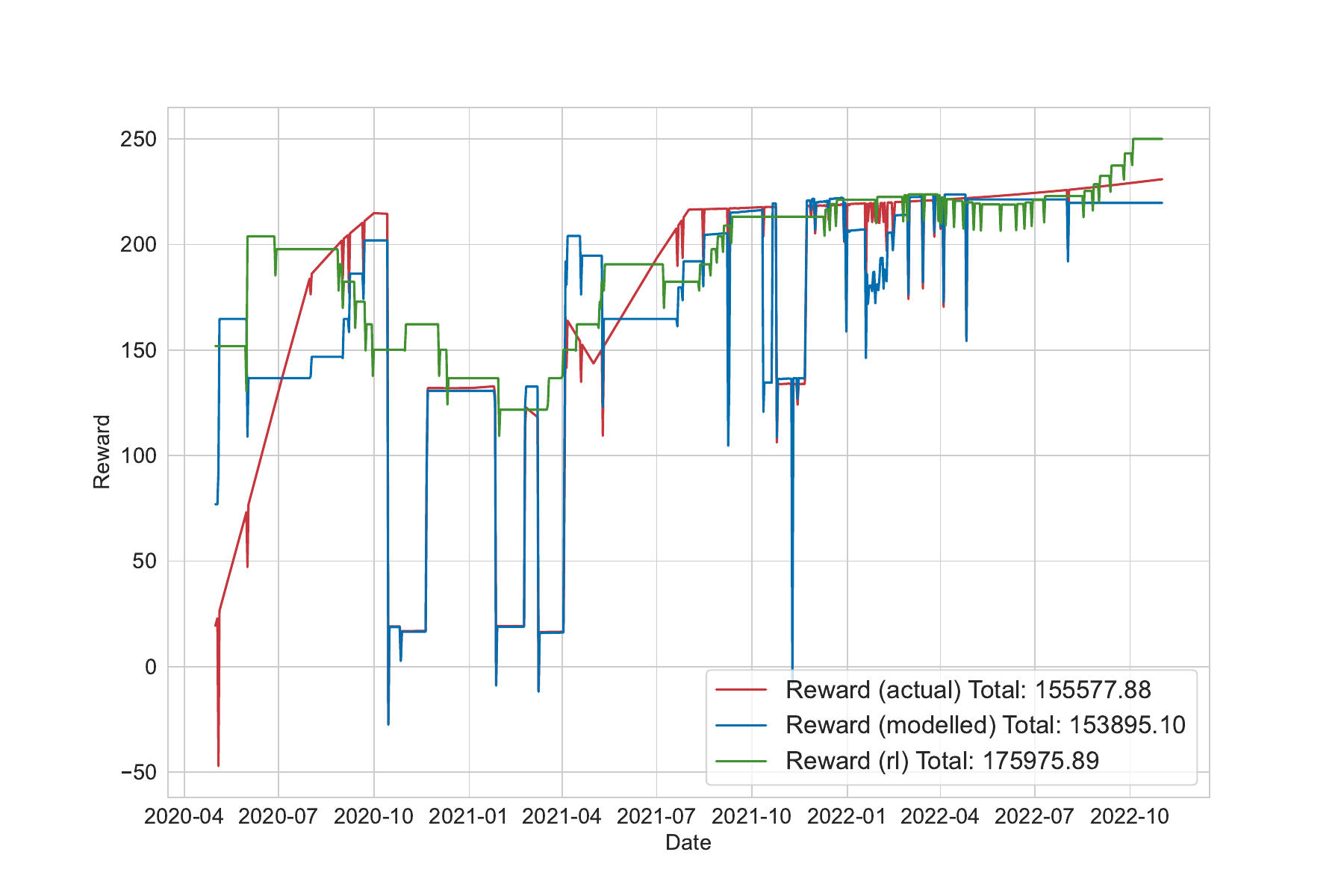}
    \caption{Reward changing over Time}
    \label{fig:175975_rl_reward}
  \end{subfigure}
  \caption{\textbf{Strategy from Reinforcement Learning Agent.} Here, "(actual)" is the real data, "(modelled)" is the result from real world stringency imposed with the use of the SIR model with lockdown and time-varying vaccination rate, and "(rl)" is the new stringency strategy we propose. \textbf{(a)} The new strategy proposed highlights an increase in stringency from October, 2020 till April, 2021 compared to the actual data and then a steady decline towards the end. \textbf{(c)} No sharp peaks in the infected population is observed. \textbf{(d)} The normalized GDP is affected by the increase in the stringency from October, 2020 till April, 2021. \textbf{(e)} The $R_e$ is maintained below $1.2$ throughout. \textbf{(f)} A higher reward is achieved by the reinforcement learning agent.}
  \label{fig:175975_parent}
\end{figure}

In contrast, an alternative output is presented in \cref{fig:175975_parent}, which demonstrates a gradual increase in stringency from October, 2020 till April, 2021. This strategy results in a decline of infections occurring prior to the vaccine's release, and a subsequent cessation of new infections. This approach, however, has implications for the normalized GDP, as seen by the observed decline in \cref{fig:175975_rl_gdp}. While both these approaches (\cref{fig:175975_parent,fig:176647_parent}) outperform the actual strategy, they underscore the complexity of managing public health crises and the need for careful strategic planning to balance health outcomes with economic considerations.

\section{Discussion}
The paper seeks to inspire epidemiologists by highlighting the advancements achieved through the application of reinforcement learning in policymaking during the pandemic. We introduce a virtual environment that closely simulates a pandemic scenario and thoroughly explore innovative strategies for disease mitigation using reinforcement learning. Our proposed approach demonstrates compelling efficacy in achieving optimal decision-making, effectively balancing the formidable challenges posed by the pandemic and economic considerations. We are confident that this research contribution will forge a connection between epidemic studies and reinforcement learning, offering valuable insights that will help humanity better defend against potential pandemic crises in the future.

\section{Experiment Settings}
\subsection{Dataset}
The population-level epidemiological data can be obtained from the "Our World In Data COVID-19" dataset: \url{https://ourworldindata.org/coronavirus} or more specifically: \url{https://github.com/owid/covid-19-data/blob/master/public/data/owid-covid-data.csv}~\cite{owidcoronavirus}. Data for the total cases, and recovered was acquired by scraping the Worldometers website~\cite{WorldometerCorona} using Internet Archive~\cite{InternetArchive}. Quaterly GDP data can be obtained from the "Organisation for Economic Co-operation and Development": \url{https://www.oecd-ilibrary.org/economics/data/main-economic-indicators/main-economic-indicators-complete-database_data-00052-en}~\cite{economic_indicators_data}.

\subsection{Code}
All code and data will be made open source upon acceptance of the paper.

\subsection{Data Availablity}
All the data used in the manuscript has been obtained from open data sources and has been cited in the manuscript itself.

\bibliography{newsample}
\end{document}